\documentclass[nohyperref]{article}

\usepackage[accepted]{icml2023}

\usepackage[textsize=tiny]{todonotes}

\usepackage{macros}

\icmltitlerunning{On the Stepwise Nature of SSL}

\begin{document}

\twocolumn[
\icmltitle{On the Stepwise Nature of Self-Supervised Learning}

\newif\ifarxiv
\arxivfalse

{
\icmlsetsymbol{equal}{*}

\begin{icmlauthorlist}
\icmlauthor{James B. Simon}{berk,gi}
\icmlauthor{Maksis Knutins}{gi}
\icmlauthor{Liu Ziyin}{tok}
\icmlauthor{Daniel Geisz}{berk}
\icmlauthor{Abraham J. Fetterman}{gi}
\icmlauthor{Joshua Albrecht}{gi}
\end{icmlauthorlist}

\icmlaffiliation{berk}{UC Berkeley}
\icmlaffiliation{gi}{Generally Intelligent}
\icmlaffiliation{tok}{University of Tokyo}

\icmlcorrespondingauthor{James Simon}{james.simon@berkeley.edu}
}

\icmlkeywords{self-supervised learning, SSL, contrastive learning, kernels, kernel methods, infinite width, NTK, deep learning, generalization, inductive bias, implicit bias, spectral bias, implicit regularization, Barlow Twins, SimCLR, CLIP}

\vskip 0.3in
]

\printAffiliationsAndNotice{}

\begin{abstract}

We present a simple picture of the training process of joint embedding self-supervised learning methods.
We find that these methods learn their high-dimensional embeddings one dimension at a time in a sequence of discrete, well-separated steps.
We arrive at this conclusion via the study of a linearized model of Barlow Twins applicable to the case in which the trained network is infinitely wide.
We solve the training dynamics of this model from small initialization, finding that the model learns the top eigenmodes of a certain contrastive kernel in a stepwise fashion, and obtain a closed-form expression for the final learned representations.
Remarkably, we then see the same stepwise learning phenomenon when training deep ResNets using the Barlow Twins, SimCLR, and VICReg losses.
Our theory suggests that, just as kernel regression can be thought of as a model of supervised learning, \textit{kernel PCA} may serve as a useful model of self-supervised learning.
\end{abstract}

\section{Introduction}
\label{sec:intro}

\begin{figure*}
  \centering
  \includegraphics[width=15cm]{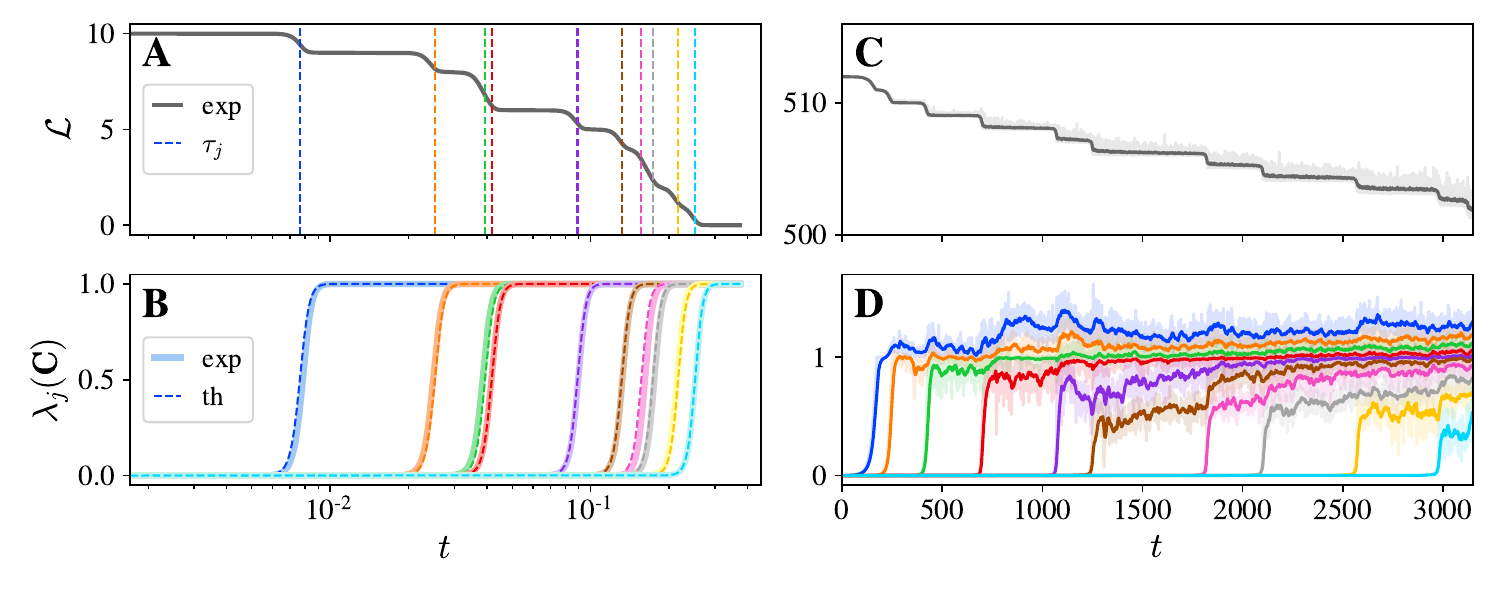}
  \vspace{-3mm}
  \caption{
    \textbf{SSL methods learn embeddings one dimension at a time in a series of discrete steps.}
    \textbf{(A,B)}: Loss and embedding eigenvalues for an analytical model of Barlow Twins using a linear model trained on $n=500$ positive pairs from CIFAR-10. Solid curves show experimental trajectories, and dashed curves show predicted step timescales (A) and eigenvalues (B).
    Curves are plotted against effective time $t = \text{[lr]}\times\text{[step]}$.
    \textbf{(C,D)}: Loss and embedding eigenvalues for Barlow Twins using a deep ResNet on STL-10 with small initialization and $\lambda=1$.
    Learning takes place in discrete, well-separated steps, each of which entails a drop in the loss and an increase in the dimensionality of the embeddings by one.
  }
  \label{fig:banner_figure}
\end{figure*}

Self-supervised learning (SSL) has recently become a leading choice for representation learning using deep neural networks.
Joint embedding methods, a prominent class of SSL methods, aim to ensure that any two ``views" of the same input --- for example, two random crops, or an image and its caption --- are assigned similar representations.
Such self-supervised approaches have yielded a bounty of recent empirical breakthroughs across domains
\citep{hjelm:2018-deep-infomax,
wu:2018-unsupervised-feature-learning,
bachman:2019-cl-from-multiple-views,
he:2020-moco,
henaff:2020-contastive-coding,
chen:2021-simsiam,
radford:2021-clip,
caron:2021-dino,
assran:2022-msn}

Despite SSL's simplicity, however, the field lacks a complete theoretical understanding of the paradigm. A promising place to start may be the fact that although there exist many different self supervised loss functions, these variants all achieve rather similar performance.\footnote{For example, \citep{bardes:2021-vicreg} report that SimCLR, SwAV \citep{caron:2020-swav}, Barlow Twins, VICReg, and BYOL \citep{grill:2020-byol} all score within a few percent of each other on ImageNet pretraining tasks.}
This similarity suggests that there may exist basic commonalities in their learning behavior which are as yet unidentified.

In this work, we propose a candidate for such a shared behavior. We present evidence that the high-dimensional embeddings of Barlow Twins \citep{zbontar:2021-barlow-twins}, SimCLR \citep{chen:2020-simclr}, and VICReg \citep{bardes:2021-vicreg} are learned one dimension at a time in a series of discrete learning stages (see Figure \ref{fig:banner_figure}). These embeddings are initially low-rank and increase in rank by one as each dimension is learned.

To reach this conclusion, we first study the Barlow Twins loss \citep{zbontar:2021-barlow-twins} applied to a linear model. The training of this model takes the form of a matrix factorization problem, and we present exact solutions to the training dynamics from small initialization. These solutions reveal that representation learning occurs in discrete, well-separated steps, each of which learns a top eigendirection of a certain contrastive kernel which characterizes the spectral bias of the model on the dataset. We also present a kernelization of our theory that allows its application to generic kernel machines, a class which importantly includes infinite-width neural networks.

Finally, we empirically examine the training of Barlow Twins, SimCLR, and VICReg using ResNets with various initializations and hyperparameters and in all cases clearly observe the stepwise behavior predicted by our analytical model.
This behavior is most apparent upon small parameterwise initialization (Figures \ref{fig:banner_figure}C,D and \ref{fig:deep_net_exps}C-F) but persists even in realistic configurations (Figure \ref{fig:dynamics_realistic}).
In light of this agreement, we suggest that SSL can be understood as a sequential process of learning orthogonal scalar functions in order of importance.
This simple picture of SSL has the potential to both provide a useful lens for understanding the training behavior of self supervised networks and help guide the design of new methods.

Concretely, our contributions are as follows:
\begin{itemize}
    \item We propose a minimal model of Barlow Twins and solve its dynamics from near-zero initialization.
    \item We extend our model to arbitrary kernel machines including infinite-width neural networks and give closed-form equations for the final embeddings in terms of the training kernel.
    \item We validate our theory using Barlow Twins, VICReg and SimCLR with deep ResNets and find good qualitative agreement.
\end{itemize}

\subsection{Motivation}

Our theoretical approach is motivated by the infinite-width ``neural tangent kernel" (NTK) limit of deep neural networks, in which the model's training dynamics become linear in its parameters \citep{jacot:2018, lee:2019-ntk}.
This limit has been immensely fruitful for the study of ordinary supervised learning with deep neural networks, yielding a variety of useful insights and intuitions regarding training and generalization, many of which hold well even for the finite, highly nonlinear models used in practice.\footnote{An incomplete list of examples includes
\citet{du:2019-convergence,
cao:2019-spectral-bias,
yang:2019-tensor-programs-I,
tancik:2020-nerf-with-fourier-features,
fort:2020-deep-learning-versus-kernel-learning,
yang:2021-tensor-programs-IV,
atanasov:2021-silent-alignment}
for training and
\citet{mei:2019-double-descent,
allen:2019-learning-in-wide-nets,
yang:2019-hypercube-spectral-bias,
arora:2019-NTK-gen-bound,
canatar:2021-spectral-bias,
simon:2021-eigenlearning,
mallinar:2022-taxonomy,
loureiro:2021-learning-curves,
wei:2022-more-than-a-toy}
for generalization.}
Kernel regression with the NTK now serves as a useful simple model for supervised deep neural networks.
Our aim here is to play the same trick with \textit{self-supervised} learning, obtaining an analytical model that can be interrogated to answer questions about SSL training.

A further motivation is the fact that deep learning methods that learn flexible \textit{representations} from unlabeled data, rather than a narrow target function from labeled data, are increasingly dominant in the modern deep learning ecosystem \citep{brown-2020-gpt3, ramesh:2022-dalle2, chowdhery:2022-palm}. Despite this, many theoretical tools have been developed exclusively in the context of supervised learning.
We aim to close this gap by porting theoretical tools from the study of supervised learning, such as kernel equivalences and spectral bias, over to the study of representation learning.

Similarly, perhaps the biggest open problem in the theory of supervised deep learning is understanding the process of \textit{feature learning} at hidden layers.
It is known that networks in the kernel limit do not adapt their hidden representations to data, so various works have explored avenues beyond the NTK, though none of these efforts have yet yielded simple closed-form equations for the features learned after training \citep{yang:2021-tensor-programs-IV, roberts:2022-principles-of-dl-theory, bordelon:2022-feature-learning-dmft, jacot:2022-dnns-with-weight-decay}.
However, we find that networks in the kernel limit \textit{do} learn nontrivial data embeddings when trained with SSL, and, what is more, we can obtain closed-form expressions for these final embeddings in terms of the NTK (Section \ref{sec:kernel_th}).
Our work presents an avenue by which feature and representation learning might be studied purely in terms of kernels.
\section{Related works}
\label{sec:related_works}

\textbf{Theoretical study of SSL.}
Many recent works have sought to understand SSL through varied lenses including statistical learning theory
\citep{arora:2019-cl-learning-theory, wei:2020-self-training-theory, nozawa:2021-understanding-negative-samples, ash:2021-investigating-negative-samples, haochen:2021-provable-guarantees-for-cl},
information theory
\citep{tsai:2020-ssl-multi-view, tsai:2021-ssl-rel-pred-coding, tosh:2021-cl-and-posterior-info, tosh:2021-cl-and-redundancy},
loss landscapes and training dynamics
\citep{tian:2020-understanding-ssl, wang:2020-understanding-cl-thru-align-and-unif, chen:2021-properties-of-contrastive-losses, tian:2021-understanding-ssl-no-neg-pairs, jing:2021-ssl-dim-collapse, wen:2021-cl-feature-learning, pokle:2022-noncontrastive-cl-landscapes, ziyin:2022-loss-landscapes, assran:2022-ssl-hidden-cluster-prior},
and kernel and spectral methods
\citep{kiani:2022-joint-embedding, haochen:2021-provable-guarantees-for-cl, johnson:2022-optimal-contrastive-reps}.
Our work unifies dynamical and kernel perspectives of SSL pioneered by prior works as follows.

\citet{ziyin:2022-loss-landscapes} showed that most common SSL loss functions take the form $\Tr{\mW \mA \mW\T} + \Tr{(\mW \mB \mW\T)^2}$ for symmetric matrices $A$ and $B$ when expanded about the origin.
This is the form of the loss function whose dynamics we solve to obtain exact optimization trajectories, and our exact solution can be adapted to other losses of the same form.

In influential work, \citet{haochen:2021-provable-guarantees-for-cl} showed that the optimal representation under a particular contrastive loss consists of the top Laplacian eigenmodes of a certain ``augmentation graph" defined over the data space.
However, as discussed by \citet{saunshi:2022-understanding-cl-req-ind-bias}, their approach is model-agnostic, and these optimal representations are usually highly degenerate in realistic cases.\footnote{Specifically, in order to be nonvacuous, their approach requires that it is common that a particular image will appear as an augmentation of \textit{multiple different base images} so that the augmentation graph is largely connected. This will virtually never happen with realistic dataset sizes and step counts.}
This degeneracy must somehow be broken by the inductive bias of the model and optimization procedure.
Our results complete this picture, characterizing this inductive bias for linearized SSL models as a bias towards small RKHS norm
and thereby identifying which of the many zero-loss solutions is reached in practice.

In a similar spirit as our work, \citet{balestriero:2022-ssl-and-spectral-methods}, \citet{kiani:2022-joint-embedding}, and \citet{cabannes:2023-ssl-interplay} study SSL with linear models and kernel methods, deriving and comparing optimal representations for certain toy losses.
They, too, find that kernelized SSL preferentially learns the top eigenmodes of certain operators, though their final representations differ from ours.
One key difference is that these works jump to optima by solving for minimizers of the loss, whereas our focus is on the dynamics of training that lead to optimal representations.
Since the loss often has many inequivalent global minimizers in practice (as is indeed the case for Barlow Twins), understanding these dynamics is necessary for determining \textit{which} global minimum will actually be found by gradient descent.

\textbf{Matrix factorization and deep linear networks.}
Our minimal model of Barlow Twins turns out to be closely related to classic matrix factorization problems \citep{gunasekar:2017-implicit-reg-in-mf, li:2018-matrix-sensing, arora:2019-implicit-reg-in-deep-mfac, chi:2019-nonconvex-opt-meets-mfac}. For example, our Equation \ref{eqn:loss_WGW} resembles Equation 3.1 of \citet{li:2018-matrix-sensing}.
Matrix factorization problems are often studied as paradigmatic examples of tasks with many inequivalent zero-loss solutions, with the challenge then to characterize the model's inductive bias and understand which of these solutions is actually reached by gradient descent.
This is often doable under an assumption of small initialization, from which gradient descent finds the solution which minimizes a certain matrix norm \citep{gunasekar:2017-implicit-reg-in-mf, li:2018-matrix-sensing}, and this is indeed the case in our model.
Our work draws a new link between SSL and matrix factorization, and we propose that degenerate matrix factorization problems can serve as illuminating simplified models for representation learning tasks more broadly.

The dynamics of our analytical model (Proposition \ref{prop:exact_dynamics}) bear a striking resemblance to the dynamics of deep linear networks.\footnote{Note that the training of a deep linear network is itself an asymmetric matrix factorization problem.} \citep{fukumizu:1998-deep-linear-nets, saxe:2013-deep-linear-nets, du:2018-deep-linear-net-optimization, arora:2018-deep-linear-net-optimization, jacot:2021-saddle-to-saddle}
Deep linear networks initialized near zero also exhibit learning in discrete, well-separated stages, each of which entails the learning of one singular direction in the model function\footnote.{Furthermore, in both cases, the loss plateaus between consecutive learning stages occur when the model passes near saddle points, with the index of the saddle point decreasing by one each stage \citep{jacot:2021-saddle-to-saddle}.}
However, the problems differ in key ways: in our setting, the stagewise learning behavior is a result of the self-supervised loss function, not the depth of the network, and our problem has many inequivalent global minima, unlike the typical deep linear setting.

\textbf{Physics and symmetry breaking.}
Interestingly, \citet{landau:1944-turbulence} encountered the same differential equation we find for embedding eigencoefficients in the study of the onset of turbuluent fluid flow.
This is a consequence of the fact that both representation learning and the onset of turbulence are processes of spontaneous symmetry breaking.
We make this connection explicitly in Appendix \ref{app:symm}.
\section{Preliminaries}
\label{sec:preliminaries}

We will study an ordinary linear model trained in a contrastive fashion.
Suppose we have a dataset of $n$ ``positive pairs" $\vx_i, \vx'_i \in \R^m$ for $i \in 1...n$.\footnote{We use a finite dataset of $n$ positive pairs for simplicity, but our theory works equally well when optimizing on the population loss of a data distribution, which simply corresponds to the case $n \rightarrow \infty$.}
Our model will consist of a linear transformation to a $d$-dimensional embedding parameterized as $\vf(\vx) \equiv \mW \mX$ with $\mW \in \R^{d \times m}$.
We would like to learn a transformation $\mW$ such that positive pairs have similar embeddings but the full set of embeddings maintains high diversity.

To encourage such representations, Barlow Twins prescribes a loss function which pushes the cross-correlation matrix between $\vf(\vx)$ and $\vf(\vx')$ towards the identity matrix.
We will use a slightly simplified variant of the Barlow Twins loss given by
\begin{equation} \label{eqn:loss_fn}
\L = \norm{\mC - \mI_d}_F^2,
\end{equation}
where $\norm{\cdot}_F$ is the Frobenius norm and $\mC \in \R^{d \times d}$ is the cross-correlation matrix given by
\begin{equation}
\mC \equiv \frac{1}{2n} \sum_i \left( \vf(\vx_i) \vf(\vx_i')\T + \vf(\vx_i') \vf(\vx_i)\T \right).
\end{equation}
Compared to the original loss of \citep{zbontar:2021-barlow-twins}, we have set the hyperparameter $\lambda$ to $1$, removed batchnorm in the definition of $\mC$, and symmetrized $\mC$.

We will initialize $\mW = \mW_0$ and train with full-batch gradient flow as
\begin{equation} \label{eqn:dW_dt}
    \frac{d\mW}{dt} = - \nabla_\mW \L.
\end{equation}
We wish to understand both the dynamics of this training trajectory and the final weights $\mW_\infty \equiv \lim_{t \rightarrow \infty} \mW$.
\section{Solving the dynamics of the linear model}
\label{sec:linear_th}

\subsection[The feature cross-correlation matrix]{The feature cross-correlation matrix $\mGamma$}
The task we have set up is a matrix optimization problem.
To elucidate the structure of this problem, we can simplify $\mC$ as
\begin{align}
    \mC &= \frac{1}{2n} \sum_i
    \mW \left( \vx_i {\vx_i'}\T \! + \vx_i' \vx_i\T \right) \mW\T = \mW \mGamma \mW\T
\end{align}
where we have defined the feature cross-correlation matrix $\mGamma \equiv \frac{1}{2n} \sum_i ( \vx_i {\vx_i'}\T \! + \vx_i' \vx_i\T )$.
Equation \ref{eqn:loss_fn} then becomes
\begin{equation} \label{eqn:loss_WGW}
    \L = \norm{\mW \mGamma \mW\T - \mI_d}_F^2,
\end{equation}
a form reminiscent of matrix factorization problems, and
Equation \ref{eqn:dW_dt} is
\begin{equation} \label{eqn:dW_dt_expanded}
\frac{d \mW}{d t} = - 4 \left( \mW \mGamma \mW\T - \mI_d \right) \mW \mGamma.
\end{equation}
We will denote by $\gamma_1 \ge \ldots \ge \gamma_m$ the eigenvalues of $\mGamma$ and, for any $k \in 1 \ldots m$, let $\mGamma^{(\le k)} \in \R^{k \times m}$ be the matrix containing the top $k$ eigenvectors of $\mGamma$ as rows.

\subsection{Exact solutions for aligned initialization}
It is nontrivial to solve Equation \ref{eqn:dW_dt_expanded} from arbitrary initialization.
However, as is common in matrix factorization problems, we can obtain exact trajectories starting from special initializations, and these special solutions will shed light on the general dynamics.
We first consider an ``aligned initialization" in which the right singular vectors of $\mW_0$ are the top eigenvectors of $\mGamma$.
Concretely, let
\begin{equation} \label{eqn:W0_spectral_def}
    \mW_0 = \mU \mS_0 \mGamma^{(\le d)}
\end{equation}
be the singular value decomposition of $\mW_0$ with $\mU \in \R^{d \times d}$ an arbitrary orthonormal matrix, and $\mS_0 \in \R^{d \times d}$ is a matrix of singular values given by
\begin{equation} \label{eqn:S0_def}
    \mS_0 = \diag(s_1(0), ..., s_d(0))
\end{equation}
with $s_j(0) > 0$.\footnote{We assume alignment with the top $d$ eigenvectors both for notational simplicity and because this is the solution we will ultimately care about, but our exact solution will hold for any set of $d$ eigenvectors of $\mGamma$.}
The dynamics of $\mW(t)$ under Equation \ref{eqn:dW_dt_expanded} are then given by the following Proposition:
\begin{proposition}[Trajectory of $\mW(t)$ from aligned initialization]
\label{prop:exact_dynamics}
If $\mW(0) = \mW_0$ as given by Equations \ref{eqn:W0_spectral_def} and \ref{eqn:S0_def}, then
\begin{equation} \label{eqn:aligned_W_decomp}
\mW(t) = \mU \mS(t) \mGamma^{(\le d)}
\end{equation}
with
$\mS(t) = \diag(s_1(t), ..., s_d(t))$
and
\begin{equation} \label{eqn:s_j(t)}
    s_j(t) = \frac{e^{4 \gamma_j t}}{\sqrt{s_j^{-2}(0) + (e^{8 \gamma_j t} - 1) \gamma_j}}.
\end{equation}
\end{proposition}

\textit{Proof of Proposition \ref{prop:exact_dynamics}}.
Treating Equation \ref{eqn:aligned_W_decomp} as an ansatz and inserting it into Equation \ref{eqn:dW_dt_expanded}, we find that
\begin{equation}
    \frac{d \mW}{d t} =
    4 \mU (1 - \mD \mS(t)^2) \mD \mS(t) \mGamma^{(\le d)},
\end{equation}
with $\mD = \text{diag}(\gamma_1, \ldots, \gamma_d)$.
It follows that the singular vectors of $\mW(t)$ remain fixed, and the singular values evolve according to
\begin{equation} \label{eqn:sv_ode}
    s_j'(t) = 4 \left( 1 - \gamma_j s_j^2(t) \right) \gamma_j s(t).
\end{equation}

This ODE can be solved to yield Equation \ref{eqn:s_j(t)}.
\QED

When $t \rightarrow \infty$, Proposition \ref{prop:exact_dynamics} prescribes singular values equal to
\begin{equation} \label{eqn:s_j_inf}
s_j(\infty) = \left\{\begin{array}{ll}
        \gamma_j^{-1/2} & \text{for } \gamma_j > 0, \\
        s_j(0) & \text{for } \gamma_j = 0, \\
        0 & \text{for } \gamma_j < 0.
        \end{array}\right.
\end{equation}
Each singular value thus flows monotonically towards either $\gamma_j^{-1/2}$ or zero depending on whether the corresponding eigenvalue is positive or negative.
This can be understood by noting that the loss (Equation \ref{eqn:loss_WGW}) can be rewritten as
\begin{equation} \label{eqn:L_as_eigensum}
    \L = \sum_j (1 - \gamma_j s_j^2)^2,
\end{equation}
which makes clear that if $\gamma_j > 0$, then $s_j = \gamma_j^{-1/2}$ is optimal (and achieves zero loss on the $j$-th term of the sum), but if $\gamma_j < 0$, then the model can do no better than $s_j = 0$

It is worth noting that $\lambda_j = \gamma_j s_j^2$ is the corresponding eigenvalue of $\mC$.
With this change of coordinates, the trajectories of Proposition \ref{prop:exact_dynamics} become nearly sigmoidal, with $\lambda_j \approx (1 + \lambda_j^{-1}(0) e^{-8 \gamma_j t})^{-1}$ when $|\lambda_j(0)| \ll 1$.

We will be particularly interested in the set of terminal solutions.
Accordingly, let us define the set of \textit{top spectral} $\mW$ as follows:
\begin{definition}(Top spectral $\mW$) \label{def:spectramax}
A top spectral $\mW$ is one for which $\mW = \mU \tilde{\mS} \mGamma^{(\le d)}$,
with $\mU$ an orthogonal matrix and
$\tilde{\mS} = \diag(\gamma_1^{-1/2} \mathbbm{1}_{\gamma_1 > 0}, ..., \gamma_d^{-1/2} \mathbbm{1}_{\gamma_d > 0})$.
\end{definition}
(Note that these are precisely the set of $\mW(\infty)$ found by Proposition \ref{prop:exact_dynamics} save for the edge case $\gamma_j = 0$, in which case we set $s_j = 0$.)
These solutions form an equivalence class parameterized by the rotation matrix $\mU$.
As observed by \citet{haochen:2021-provable-guarantees-for-cl}, such a rotation makes no difference for the downstream generalization of a linear probe, so we may indeed view all top spectral $\mW$ as equivalent.

Let us assume henceforth that $\gamma_1, ..., \gamma_d > 0$ and $\gamma_d > \gamma_{d+1}$.
The top spectral $\mW$ achieve $\L(\mW)=0$, but note that there generally exist other optima, such as those aligned with a different set of positive eigenvectors.
However, the top spectral $\mW$ are optimal in the following sense:
\begin{proposition} \label{prop:Fnorm_min}
The top spectral $\mW$ are precisely the solutions to
\begin{equation}
\underset{\mW}{\text{argmin}} \norm{\mW}_F
\ \ \
\text{s.t.}
\ \ \
\L(\mW) = 0.
\end{equation}
\end{proposition}
We relegate the proof to Appendix \ref{app:derivation}.
Proposition \ref{prop:Fnorm_min} implies that, of all solutions achieving $\L = 0$, the top spectral solutions have minimal $\norm{\mW}_F$.
Noting that gradient descent often has an implicit bias towards low-norm solutions \citep{gunasekar:2017-implicit-reg-in-mf, soudry:2018-implicit-bias-of-gd}, we might expect to reach this set of optima from some more general initial conditions.

\subsection{The case of small initialization}
\label{subsec:small_init}
Returning to Proposition \ref{prop:exact_dynamics}, an informative special case of our exact dynamical solution is that in which the initial singular values are small relative to their final values $(s_j(0) \ll \gamma_j^{-1/2})$.
In this case, Equation \ref{eqn:s_j(t)} states that $s_j(t)$ will remain small up to a critical time
\begin{equation} \label{eqn:tau_j_def}
\tau_j = \frac{- \log (s_j^2(0) \gamma_j)}{8 \gamma_j}
\end{equation}
after which it will rapidly grow to its final value.
Note that $\tau_j$ is principally set by $\gamma_j$
, with only a weak logarithmic dependence on initialization $s_j(0)$.
The learning dynamics can thus be understood as a \textit{stepwise process} in which $d$ orthogonal directions are each rapidly learned at their respective timescales $\tau_j$, with plateaus in between adjacent learning phases.

Proposition \ref{prop:exact_dynamics} assumed a special aligned initialization for $\mW$.
We will now give a result which generalizes this significantly, showing that the trajectory from any \textit{small} initialization closely follows that from a particular aligned initialization.

In order to state our result, we will first define the QR factorization and an ``alignment transformation."
\begin{definition}[QR factorization.]
The \textit{QR factorization} of a matrix $\mM \in \R^{a \times b}$ returns a decomposition $\mQ \mR = \mM$ such that $\mQ \in \R^{a \times a}$ is orthogonal and $\mR \in \R^{a \times b}$ is upper-triangular with nonnegative diagonal.
If $a \le b$ and $\mM$ is full rank, then the QR factorization is unique.
\end{definition}
\begin{definition}[Alignment transformation.]
The \textit{alignment transformation} $\mathcal{A}$ of a matrix $\mM$ returns a matrix $\mathcal{A}(\mM) = \mQ \tilde{\mR}$, where $\mQ\mR = \mM$ is a QR factorization and $\tilde\mR$ is $\mR$ with all off-diagonal elements set to zero.
\end{definition}

We can now state the main result of this section.
\begin{result}[Trajectory from generic small initialization] $ $
\label{res:small_init}
\vspace{-2mm}
\begin{itemize}
    \item Let $\gamma_1, ..., \gamma_d$ be unique.
    \item Let $\tilde{\mW}_0 \in \R^{d \times m}$ with $\tilde{\mW}_0 \mGamma^{(\le d)}$ full rank.
    \item Let $\mW(t)$ be the solution to Equation \ref{eqn:dW_dt_expanded} with initial condition $\mW(0) = \alpha \tilde\mW_0$.
    \item Let $\mW^*(t)$ be the aligned solution with initial condition $\mW^*(0) = \mathcal{A}(\mW(0) {\mGamma^{(\le m)}}\T) \mGamma^{(\le m)}$.
\end{itemize}
Then as $\alpha \rightarrow 0$, $\norm{\mW(t) - \mW^*(t)}_F \rightarrow 0$ for all $t$.
\end{result}
We give a derivation of this result in Appendix \ref{app:derivation}.\footnote{We style this conclusion as a Result rather than a Theorem because we give an informal derivation rather than a formal proof.
We conjecture that this result can indeed be made formal.}

Result \ref{res:small_init} states that the trajectory from generic small initialization closely follows a particular aligned trajectory.
This aligned trajectory is given in closed form by Proposition \ref{prop:exact_dynamics}, and so this result gives us equations for the dynamics from arbitrary initialization.

Some intuition for this result can be gained by examining the construction of $\mW^*(0)$.
The aligned solution $\mW^*(t) = \mU \mS^*(t) \mGamma^{(\le d)}$ is composed solely of the top $d$ eigendirections of $\mGamma$, but an arbitrary initialization will have no such preference.
How does this alignment occur?
Note that, at early times when $\mW$ is small, the quadratic term of the loss will dominate, and Equation \ref{eqn:dW_dt_expanded} reduces to
\begin{equation}
    \frac{d \mW}{d t} \approx 4 \mW \mGamma
    \ \ \ \Rightarrow \ \ \
    \mW \approx \mW(0) \, e^{4 \mGamma t}.
\end{equation}
The top eigendirections of $\mGamma$ grow faster and will quickly dominate, and after a time $\tilde{\tau} \gg (\gamma_d - \gamma_{d+1})^{-1}$, we will have
\begin{equation}
    \mW \approx \mW(0) \mPi^{(\le d)} \, e^{4 \mGamma t}
\end{equation}
where $\mPi^{(\le d)} \equiv {\mGamma^{(\le d)}}\T \mGamma^{(\le d)}$ is the projector onto the top-$d$ eigenspace of $\mGamma$.
Components aligned with eigenmodes of index $j > d$ are thus negligible compared to those of index $\le d$ and do not interfere in the learning dynamics, which converge before such later eigenmodes can grow to order one.

Having identified the relevant eigendirections, we must now determine their effective initial singular values $s^*_j(0)$.
Let $\vv_j$ be the $j$-th eigenvector of $\mGamma$ with $\norm{\vv_j} = 1$ and define $\vu_j = \mW(0) \vv_j$.
If $\vv_j$ were a right singular vector of $\mW(0)$, we would have $s^*_j(0) = \norm{\vu_j}$.
We will not be so fortunate in general, however.
Examining Equation \ref{eqn:dW_dt_expanded}, we should expect each eigenmode to only be able to grow in the subspace of $\R^d$ which has not already been filled by earlier eigenmodes, which suggests that we take
\begin{equation} \label{eqn:s_j(0)_approx}
    s^*_j(0) = \norm{\left( 1 - \sum_{k < j} \frac{\vu_k\T \vu_k}{\norm{\vu_k}^2} \right) \vu_j}.
\end{equation}
These are precisely the singular values of $\mW^*(0)$.\footnote{This can be seen by noting that the QR factorization involves the Gram-Schmidt-like process of Equation \ref{eqn:s_j(0)_approx}.}

\subsection{Numerical simulation}
\label{subsec:linear_model_sim}

We perform basic numerical simulations of our linear problem which verify Proposition \ref{prop:exact_dynamics} and Result \ref{res:small_init}.
We sample $n=500$ random images from CIFAR-10 \citep{krizhevsky:2009} and, for each, take two random crops to size $20 \times 20 \times 3$ to obtain $n$ positive pairs (which thus have feature dimension $m = 1200$).
We then randomly initialize a linear model with output dimension $d=10$ and weights drawn i.i.d. from $\mathcal{N}(0,\alpha^2)$ with $\alpha = 10^{-7}$ and train with the loss of Equation \ref{eqn:loss_fn} with learning rate $\eta = 5 \times 10^{-5}$.

During training, we track both the loss and the eigenvalues of the embedding cross-correlation matrix $\mC$.
Our stepwise dynamics predict the loss will start at $\L(0) \approx d$ and decrease by one as each mode is learned, giving
\begin{equation}
    \L(t) \approx \sum_{j | \tau_j > t} 1.
\end{equation}
The eigenvalues of $\mC$ will be $(\gamma_1 s_1^2(t), ..., \gamma_d s_d^2(t))$, with $s_j(t)$ given by Proposition \ref{prop:exact_dynamics}.
The use of Proposition \ref{prop:exact_dynamics} requires values for $s_1(0), ..., s_d(0)$, and these can be found from Equation \ref{eqn:s_j(0)_approx} and the statistics of the random initialization to be roughly
\begin{equation}
    s_j(0) \approx \sigma \sqrt{d - j + 1}.
\end{equation}
The results are plotted in Figure \ref{fig:banner_figure}(A,B).
We find excellent agreement with our theory.

\section{Kernelizing the linear model}
\label{sec:kernel_th}

Our discussion has so far dealt with an \textit{explicit} linear regression problem in which we have direct access to the data features $\vx_i$.
We used this explicit representation to construct $\mGamma \in \R^{m \times m}$, which lives in feature space.
However, many models of interest are linear with an \textit{implicit} feature space, with the output a linear function not of the input $x$ but rather of a fixed transformation of the input $\vphi(x)$.
Models of this class include kernel machines \citep{shawe:2004}, random feature models and deep neural networks in which only the last layer is trained \citep{rahimi:2007, lee:2018-nngp}, and infinite-width neural networks, the latter of which evolve under the NTK \citep{jacot:2018}.
While these models are equivalent to linear models, we do not have an explicit representation of the features $\vphi(x)$ (which may be large or even infinite-dimensional).
What we \textit{do} have access to is the model's \textit{kernel function} $\K(x,x') = \vphi(x)\T \vphi(x')$.

In our original linear model, the kernel is simply the inner product $\vx\T \vx'$.
Reinterpreting $\vx$ as $\vphi(x)$, any quantity expressed solely in terms of such inner products can still be computed after kernelization.
Our challenge, then, is to rewrite our theory of dynamics entirely in terms of such inner products so we may apply it to general kernel machines.

\subsection{Kernelized solution}

We will manage to do so.
Let our datasets be $\X \equiv \{x_1, ..., x_n\}$ and $\X' \equiv \{x'_1, ..., x'_n\}$.
Let us define the kernel matrix $\Kaa \in \R^{n \times n}$ such that $[\Kaa]_{ij} = \K(x_i,x_j)$ and define $\Kab$, $\Kba$, and $\Kbb$ analogously.
Let us also define
\begin{equation}
\tilde{\mK} \equiv \begin{bmatrix}
\Kaa & \Kab\\
\Kba & \Kbb
\end{bmatrix},
\end{equation}
the kernel over the combined dataset, as well as
\begin{equation}
\!\!\!
\mZ \equiv
\frac{1}{2n}
\begin{bmatrix}
\Kab\Kaa & \Kab^2\\
\Kbb\Kaa & \Kbb\Kab 
\end{bmatrix} + \text{[transpose]}.
\end{equation}
Finally, let us define
$\mK_\mGamma \equiv \tilde{\mK}^{-1/2} \mZ \tilde{\mK}^{-1/2} \in \R^{2n \times 2n}$,
where $\tilde{\mK}^{-1/2}$ is interpreted as $(\tilde{\mK}^+)^{-1/2}$ if $\tilde{\mK}$ is degenerate.
The matrix $\mK_\mGamma$ is symmetric and akin to a kernelized version of $\mGamma$.\footnote{Here $\mGamma = \frac{1}{2n} \sum_i \left(\vphi(x_i) \vphi(x'_i)\T + \vphi(x'_i) \vphi(x_i)\T\right)$ since we have reinterpreted $\vx$ as $\vphi(x)$. This $\mGamma$ would be sufficient to use our theory of Section \ref{sec:linear_th} were it not constructed of \textit{outer} products of $\vphi(x_i)$'s, which are inaccessible after kernelization.}\footnote{While $\mK_\mGamma$ is not technically a kernel matrix as it can have negative eigenvalues, we may heuristically think of it as one since only the (top) positive eigenvalues typically matter.}

Let $(\gamma_j, \vb_j)$ be the eigenvalues and normalized eigenvectors of $\mK_\mGamma$ indexed in descending eigenvalue order.
The kernelization of our theory is given by the following proposition.

\begin{proposition}[Kernelized solution] \label{prop:kernelized_sol} $ $
\begin{enumerate}[(a)]
\item All nonzero eigenvalues of $\mK_\mGamma$ are eigenvalues of
$\mGamma$
and vice versa.
\item \label{propclause:kernelized_embs}
The top spectral solution gives the embeddings
\begin{equation} \label{eqn:kernelized_top_spectral_embs}
    \vf(x) = \mU \tilde{\mS} [\vb_1 \ ... \ \vb_d]\T \tilde{\mK}^{-1/2} [\mK_{x \X} \ \mK_{x \X'}]\T
\end{equation}
with
$\mU$ an orthogonal matrix,
$\tilde{\mS} = \diag(\gamma_1^{-1/2}, ..., \gamma_d^{-1/2})$,
and
$\mK_{x \X}, \mK_{x \X'} \in \R^{1 \times n}$
such that 
$[\mK_{\X x}]_i = \K(x_i,x)$ and $[\mK_{\X' x}]_i = \K(x_i',x)$.
\item The top spectral solutions correspond to the embeddings $\vf(x)$ given by
\begin{equation}
    \underset{\vf}{\text{argmin}} \norm{\vf}_\K
    \ \ \ \text{s.t.} \ \ \
    \L(\vf) = 0.
\end{equation}
\end{enumerate}
\end{proposition}
We give the proof and an expression for $\vf(x,t)$, the embeddings over time from small (aligned) initialization, in Appendix \ref{app:proofs}.
These results allow us to predict both the training dynamics and final embeddings of our contrastive learning problem with a black-box kernel method.

\subsection{Implications of kernelized solution}
This kernelized solution has several interesting properties that we will briefly discuss here.

\textbf{Special case: $\X = \X'$.}
In the case in which the two views of the data are identical, one finds that $\mK_\mGamma = \tilde\mK$ and the model simply learns the top eigenmodes of its base (neural tangent) kernel.

\textbf{SSL as kernel PCA.}
Proposition \ref{prop:kernelized_sol}\ref{propclause:kernelized_embs} states that the final embeddings are governed by the top $d$ eigenvectors of the kernel-like matrix $\mK_\Gamma$.
With our setup, then,
\textbf{SSL with neural networks amounts to \textit{kernel PCA} in the infinite-width limit.}
This is analogous to the fact that standard supervised learning approaches kernel \textit{regression} in the infinite-width limit.\footnote{This analogy can be furthered by noting that these equivalences both occur as $t \rightarrow \infty$ and both require small or zero initialization.}
This is rather satisfying in light of the fact that kernel regression and kernel PCA are the simplest supervised and unsupervised kernel methods, respectively.

\textbf{The same theory works for multimodal SSL.}
The above solution holds for the infinite-width NTK limit of any neural network architecture.
It is worth noting that this includes multimodal setups such as CLIP \citep{radford:2021-clip} in which representations for the two datasets $\X$ and $\X'$ are generated by \textit{different} models, and the two datasets may be of different types.
We can view the two models as one combined model with two pathways, and since these pathways share no parameters, we have $\Kab = \Kba = \mathbf{0}$.\footnote{In the parlance of the original linear model, the feature vectors of $\X$ and $\X'$ lie in orthogonal subspaces, and the model is tasked with discovering correlations across these subspaces.}
Both our linear and kernelized solutions remain valid and nonvacuous in this setting.

\textbf{Generalization on downstream tasks.}
The quality of a learned representation is often assessed by fitting a downstream function $g^*$ (such as an image classification) with a linear function of the representation as $\hat{g}(x) = \vbeta\T \vf(x)$.
Downstream task performance will be good if $g^*$ lies largely in the linear span of the components of $\vf(x)$.
Since Proposition \ref{prop:kernelized_sol}\ref{propclause:kernelized_embs} yields $\vf(x)$ in closed form, generalization can thus be assessed in our setting.
We leave this direction for future work.

\textbf{Mapping from initial to final kernel.}
Downstream task performance will be determined by the learned kernel
$\K_{\text{emb}}(x,x') \equiv \vf(x)\T \vf(x')$,
which contains all the information of $\vf$ save for the arbitrary rotation $\mU$.
We can thus think of SSL as a process which maps an initial, naive kernel $\K$ to a final kernel $\K_{\text{emb}}$ which has learned the structure of the data.
Many other poorly-understood processes in deep learning --- most notably that of feature learning --- also have the type signature \texttt{(initial kernel, data) $\rightarrow$ (final kernel)}, but there exist few closed-form algorithms with this structure.
While representation learning and supervised feature learning are different processes, it seems likely that they will prove to be related,\footnote{See \citet{geirhos:2020-ssl-vs-sl} and \citet{grigg:2021-ssl-vs-sl} for evidence in this direction.} and thus our closed-form solution for the final kernel may be useful for the study of feature learning.

\section{Experiments}
\label{sec:exp}

\begin{figure*}
  \centering
  \includegraphics[width=17cm]{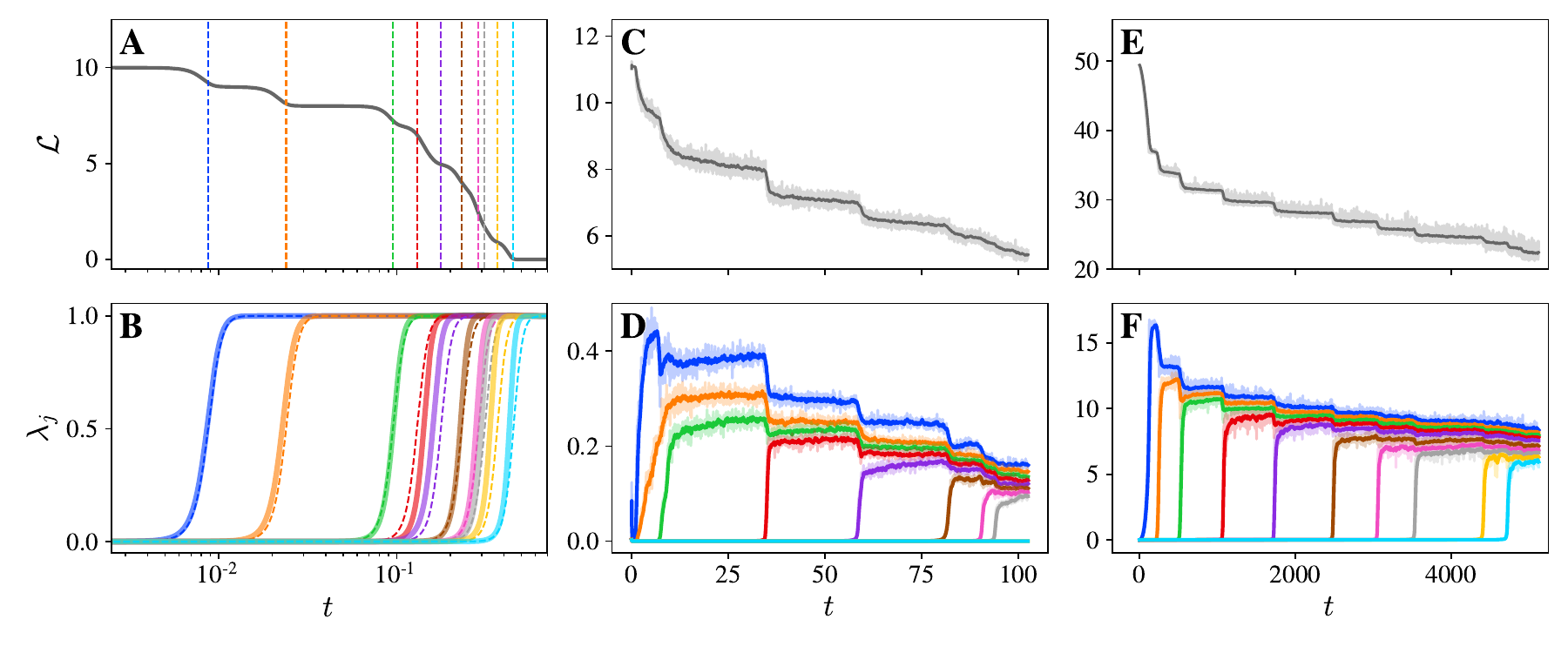}
  \vspace{-3mm}
  \caption{
    \textbf{Stepwise learning in SSL with nonlinear neural networks trained from small initialization.}
    Losses and embedding eigenvalues as a function of time $t = \text{[lr]}\times\text{[step]}$ for
    \textbf{(A, B)} a single-hidden-layer MLP trained with our simplified Barlow Twins loss,
    \textbf{(C, D)} a ResNet trained with SimCLR loss, and
    \textbf{(E, F)} a ResNet trained with VICReg loss, all trained on STL-10.
  }
  \label{fig:deep_net_exps}
\end{figure*}

Since our theory was derived in the case of linear models, a natural question is whether it is useful and informative even for the study of practical deep neural networks.
\textbf{Here we present evidence that the stepwise learning phenomenon we identified occurs even in realistic SSL setups with ResNet-50 encoders, and even with losses besides Barlow Twins.}
We sketch experiments here and provide more details in Appendix \ref{app:exp}.
In all figures, reported eigenvalues are those of the cross-correlation matrix $\mC$ when the loss is Barlow Twins, and are those of the ordinary covariance matrix of the embeddings when the loss is SimCLR or VICReg unless otherwise stated, though in practice these give quite similar results.

We perform a series of experiments with increasingly realistic models.
As a first nonlinear experiment, we revisit our previous numerical simulation of our linear model (Section \ref{subsec:linear_model_sim}) and simply replace the model with a width-2048 \text{ReLU} MLP with a single hidden layer using the standard PyTorch parameterization.
This model is fairly wide, and so we expect its NTK to remain roughly fixed.
Results are shown in Figure \ref{fig:deep_net_exps} (A,B).

We next train practical deep SSL methods with some hyperparameters modified slightly so as to better align with our theory.
We train VICReg, SimCLR, and Barlow Twins using a ResNet-50 encoder and an MLP projection head on the full STL-10 dataset \citep{coates:2011-stl10}.
These runs use small parameterwise initialization and small learning rate, but important hyperparameters such as momentum and weight decay are otherwise kept largely as in the original publications\footnote{We also changed batch size for practical reasons related to calculating the kernel and set $\lambda=1$ for Barlow Twins. See Appendix \ref{app:exp} for full details.}.
The results, plotted in Figure \ref{fig:banner_figure}(C,D) (Barlow Twins) and Figure \ref{fig:deep_net_exps}(C-F) (SimCLR and VICReg), clearly show the same stepwise learning behavior seen in simpler settings\footnote{Unlike with the three reported methods, we did not immediately see stepwise learning with BYOL.}.
Agreement with predictions using the initial NTK is poor as expected from work on ResNet NTKs \citep{fort:2020-deep-learning-versus-kernel-learning, vyas:2022-limitations-of-the-ntk-for-generalization} and we omit theory curves.

Finally, we repeat these experiments in fully realistic settings, with standard initialization and learning rates.
Even though eigenvalues are no longer uniformly small at initialization, we still see stepwise learning in their growth, as shown in Figure \ref{fig:dynamics_realistic}. 
Specifically, we see for each loss that eigenvalues separate clearly into a band of eigenvalues that have not yet grown and a band of those that have fully grown with a sparse region in between, and eigenvalues move from the lower band to the upper band as training proceeds.
This interpretation is affirmed by eigenvalue histograms throughout training, which reveal bimodal distributions as shown in Figure \ref{fig:histograms_realistic}.
Our view is that stepwise learning --- i.e., the sequential growth of the rank of the embeddings, corresponding to the learning of orthogonal functions --- is generic, and is simply made cleaner upon taking small initialization\footnote{Figure \ref{fig:dynamics_realistic} suggests that, at least for Barlow Twins and VICReg, embedding directions may be learned ``a few at a time" rather than one at a time with realistic hyperparameters, but when the embedding dimension is also realistically large, this difference is minor.}.

\textbf{Stepwise behavior in hidden representations.}
All theory and experiments thus far have studied the final embeddings of the network.
However, in practice, one typically uses the hidden representations of the network taken several layers before the output for downstream tasks.
In light of this, we repeat our experiments with small initialization but measure eigenvalues computed from the hidden representations.
Results are reported in Figure \ref{fig:rep_vals}.
Remarkably, despite the fact that our theory says nothing about hidden representations, \textit{we still clearly see learning steps coinciding with the learning steps at the embeddings.}
Understanding the couplings in these dynamics is an interesting direction for future work.

\textbf{Theoretical predictions of embeddings with the empirical NTK.}
Our theory predicts not only the occurrence of learning steps in the model embeddings but also the precise embedding function learned (up to an orthogonal transformation) in terms of the model's NTK.
As a final experiment, we compare these predictions, evaluated using the empirical NTK after training, with the true embeddings learned in ResNet-50 experiments from small initialization with $d=50$.
The subspaces spanned by the empirical and predicted embeddings have alignment significantly above chance, with the theoretical predictions capturing a bit over 50\% of the span of the true embeddings in a precise sense we define.

Intriguingly, we also find that the embeddings have comparable agreement \textit{between} SSL methods.
This suggests that these methods are in fact learning similar things, which is encouraging for the prospects for developing unifying theory for SSL.
We give details of these experiments in Appendix \ref{app:emb_pred}.
\vspace{-2mm}
\section{Conclusions}
\label{sec:conc}

We have presented and solved a linearized minimal model of Barlow Twins (Sections \ref{sec:linear_th} and \ref{sec:kernel_th}).
Our solution reveals that, as training proceeds, the model sequentially learns the top $d$ eigenmodes of a certain kernel in a discrete fashion, stopping when the embeddings are full rank.
Turning to bona fide ResNets in near-realistic training settings, we see precisely this learning behavior for Barlow Twins, SimCLR, and VICReg in both their embeddings and hidden representations.
This paints a new, useful picture of the training dynamics of SSL: rather than a black-box optimization process that magically converges on final representations, we can perhaps think of self-supervised training as an iterative process of selecting desirable rank-one functions and tacking them onto a growing representation.

Our theory has several clear limitations.
First, practical deep neural networks are \textit{not} kernel methods: the NTK is known to change over time \citep{yang:2021-tensor-programs-IV, vyas:2022-limitations-of-the-ntk-for-generalization}.
This suggests that our theory's predicted representations will not closely match those predicted by the initial NTK in practical cases, though it is plausible that they \textit{will} match those predicted by the empirical NTK \textit{after} training in light of similar results for supervised learning \citep{long:2021-after-kernel, atanasov:2021-silent-alignment}.
Second, in practice, downstream tasks in self-supervised pipelines are usually trained not on the final embeddings of the SSL model but rather on the hidden representation some layers prior, while our theory (like virtually all current SSL theory) only describes the embeddings.
We partially surmount this limitation by empirically observing stepwise behavior in representations, though developing theory for this observation appears a worthwhile pursuit.
We note that it is not currently known why the use of hidden representations is preferable, but having a theory of final embeddings like that we have given may aid efforts to understand how they differ from hidden representations.

This work opens new avenues of research from both theoretical and empirical angles.
For theory, we draw new connections between SSL, matrix factorization, and questions of inductive bias which admit further study.
Empirically, it seems plausible that our a picture of SSL learning can enable algorithmic improvements that give faster, more robust, or better-generalizing training.
The prospects for accelerating the training of SSL methods, which typically require many more steps to converge than standard supervised methods, seem particularly promising.
For example, our experiments suggest that this slow training may be due to the long times required for lower-eigenvalue modes to emerge, and that an optimizer or loss function that focuses updates on near-zero eigendirections in the embedding space may speed up training without sacrificing stability or generalization.
We describe several potential paths for realizing this speedup in Appendix \ref{app:speedup} and encourage practitioners to explore their implementation.

The clear occurrence of stepwise learning far outside the NTK regime suggests it ought to be derivable from a far less restrictive set of assumptions on the model.
We leave the development of a more generic theory for future work.
A promising starting point may be the view of stepwise learning as a consequence of symmetry-breaking which is discussed in Appendix \ref{app:symm}.

Another interesting direction is the observation of stepwise behavior in masked-image modeling frameworks, which currently constitute a large fraction of the SSL literature \citep{baevski:2022-data2vec, he:2022-mae, assran:2023-i-jepa}.
We suspect that, upon small initialization, their hidden representations may also exhibit stepwise dimensionality growth as the system learns to pass more and more information from input to output.

\vspace{-3mm}
\section*{Acknowledgements}
\vspace{-2mm}
The authors thank Bobak Kiani, Randall Balestreiro, Vivien Cabannes, Kanjun Qiu, Ellie Kitanidis, Bryden Fogelman, Bartosz Wr\'{o}blewski, Nicole Seo, Nikhil Vyas, and Michael DeWeese for useful discussions and comments on the manuscript. JS gratefully acknowledges support from the National Science Foundation Graduate Fellow Research Program (NSF-GRFP) under grant DGE 1752814.

\bibliography{ml_refs}

\bibliographystyle{icml2023}

\appendix
\onecolumn

\section{Additional figures}
\label{sec:addnl_figs}

\begin{figure}[H]
  \includegraphics[width=17cm]{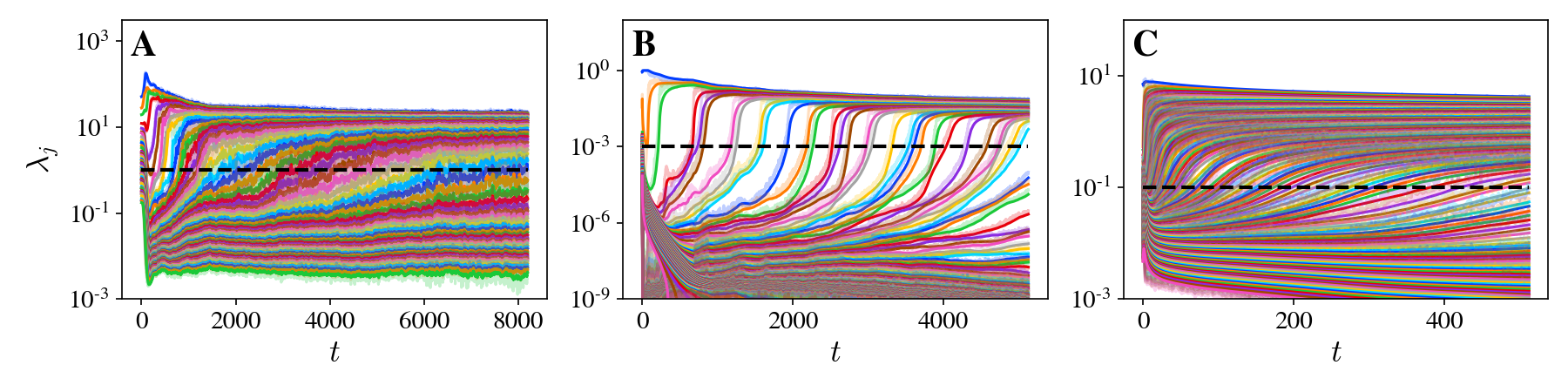}
  \vspace{-3mm}
  \caption{
    \textbf{Stepwise learning in SSL with ResNets with standard initialization and hyperparameters.}
    Embedding eigenvalues over time for \textbf{(A)} Barlow Twins, \textbf{(B)} SimCLR, and \textbf{(C)} VICReg.
    Dashed horizontal lines show rough eigenvalue thresholds separating the cluster of modes which have grown from the cluster of modes which have not yet grown and match the thresholds of Figure \ref{fig:histograms_realistic}.
  }
  \label{fig:dynamics_realistic}
\end{figure}

\begin{figure}[H]
  \includegraphics[width=17cm]{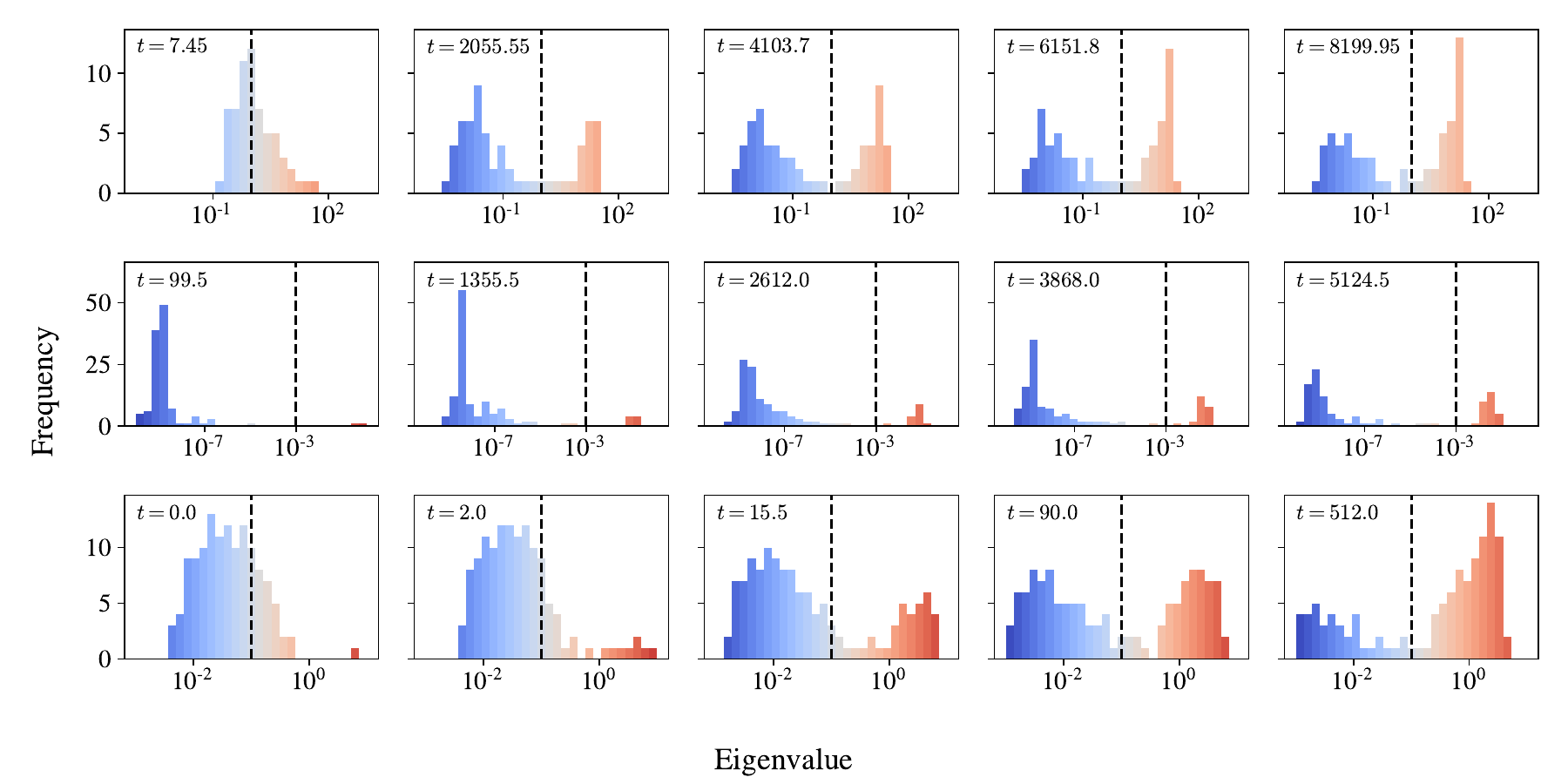}
  \vspace{-3mm}
  \caption{
    \textbf{Bimodal distribution of embedding eigenvalues shown over time for SSL with ResNets with standard initialization and hyperparameters.}.
    Histograms of embedding eigenvalues at selected times throughout training for \textbf{Top row:} Barlow Twins. \textbf{Middle row:} SimCLR. \textbf{Bottom row:} VICReg.
    Dashed vertical lines indicate the same eigenvalue thresholds as in Figure \ref{fig:dynamics_realistic}.
  }
  \label{fig:histograms_realistic}
\end{figure}

\begin{figure}[H]
  \centering
  \includegraphics[width=17cm]{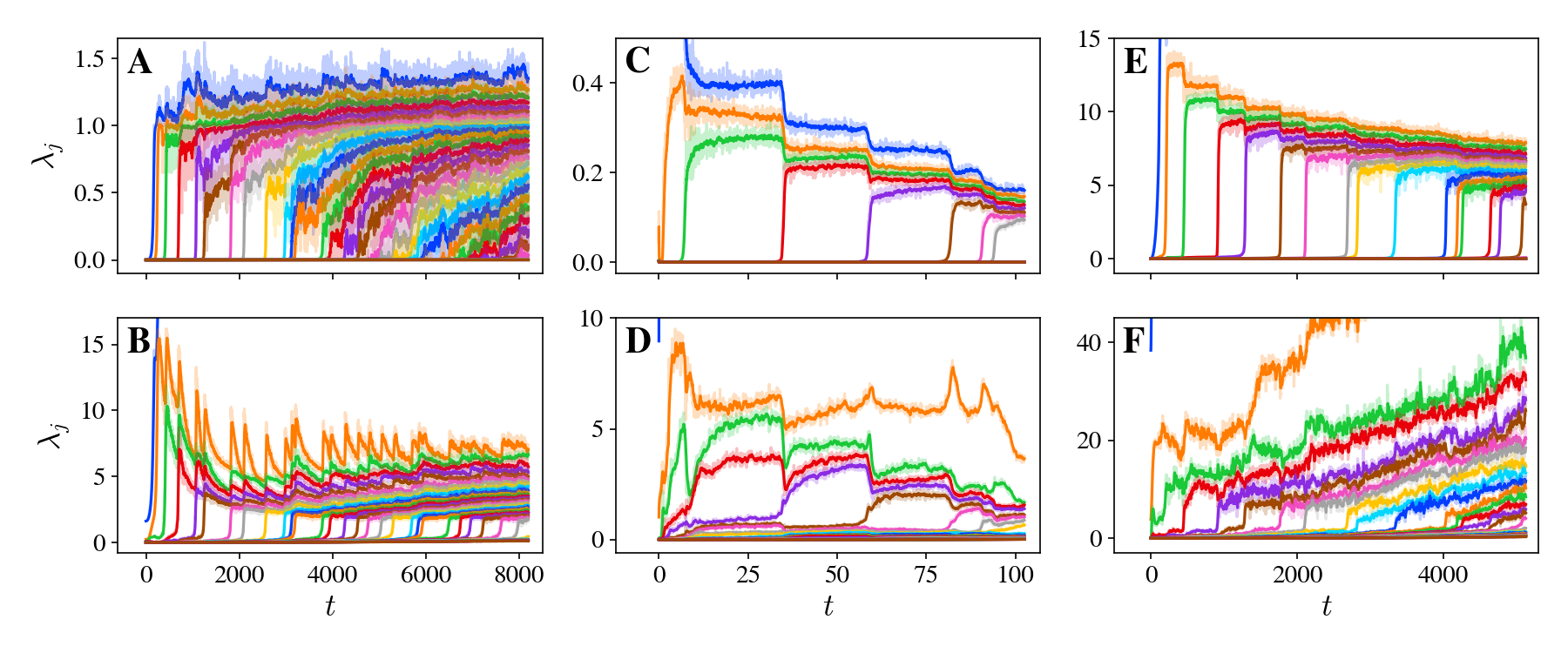}
  \vspace{-3mm}
  \caption{
    \textbf{Hidden representations exhibit learning steps aligned with those of embeddings.}
    Eigenvalues of embeddings (top row) and hidden representations (bottom row) vs time $t = \text{[lr]}\times\text{[step]}$ for
    \textbf{(A, B)} a ResNet trained with Barlow Twins loss,
    \textbf{(C, D)} a ResNet trained with SimCLR loss, and
    \textbf{(E, F)} a ResNet trained with VICReg loss, all trained on STL-10 with small initialization.
  }
  \label{fig:rep_vals}
\end{figure}
\section{Proofs}
\label{app:proofs}

\subsection{Proof of Proposition \ref{prop:Fnorm_min}}

We wish to show that the top spectral $\mW$ are the unique solutions to
\begin{equation}
    \underset{\mW}{\text{argmin}} \norm{\mW}_F
    \ \ \ \text{s.t.} \ \ \
    \L(\mW) = 0.
\end{equation}
Any such solution is also a minimum of
\begin{equation}
    \L_\epsilon(\mW) \equiv \L(\mW) + \norm{\mW}_F^2
\end{equation}
as $\epsilon \rightarrow 0$.
We will show that the set of minima are precisely the top spectral embeddings.

Any local minimum must satisfy
\begin{align}
    \nabla_\mW \L_\epsilon
    &= (\mI - \mW \mGamma \mW\T) \mW \mGamma - \epsilon \mW = \mathbf{0} \\
    \label{eqn:mW_local_min_eqn}
    \Rightarrow
    \mW\T \nabla_\mW \L_\epsilon
    &= \mW\T\mW\mGamma - \left(\mW\T\mW\mGamma\right)^2 - \epsilon \mW\T\mW = \mathbf{0}.
\end{align}
Note that the first and second terms of Equation \ref{eqn:mW_local_min_eqn} share all eigenvectors (i.e. commute), and thus the third term must commute with both of them.
From the fact that $\mW\T\mW\mGamma$ and $\mW\T\mW$ share eigenvectors, we can conclude that $\mW\T\mW$ and $\mGamma$ share eigenvectors.
The eigenvectors of $\mW\T\mW$ are simply the right singular vectors of $\mW$.

Any local minimum must thus take the form
\begin{equation}
    \mW = \mU \mS \mGamma^{(\mathcal{J})},
\end{equation}
where
\begin{itemize}
\item $\mU \in \R^{d \times d}$ is orthogonal,
\item $\mS \in \R^{d \times d} = \text{diag}(s_1, ..., s_d)$,
\item $\mathcal{J}$ is a list of $d$ elements from $\{1,...,d\}$,
\item $\mGamma^{(\mathcal{J})} \in \R^{d \times m}$ contains as rows the $d$ eigenvalues of $\mGamma$ corresponding to the elements of $\mathcal{J}$, and
\item the singular values satisfy
\begin{equation}
    s_j^2 \gamma_{\mathcal{J}_j} - s_j^4 \gamma_{\mathcal{J}_j}^2 - \epsilon s_j^2 = 0
    \ \ \ \Rightarrow \ \ \
    s_j \in \left\{0, \sqrt{\frac{\gamma_{\mathcal{J}_j} - \epsilon}{\gamma_{\mathcal{J}_j}^2}}\right\},
\end{equation}
where $s_j = 0$ is chosen by default when $\gamma_{\mathcal{J}_j} \le 0$.
\end{itemize}
Of these candidate local minima, a global minimum must choose $s_j = 0$ as infrequently as possible, and given that must minimize $\norm{\mW}_F^2 = \sum_j s_j^2$.
This is achieved by choosing $\mathcal{J} = (1,...,d)$, which yields the top spectral solutions.

\subsection{Proof of Proposition \ref{prop:kernelized_sol}}
\newcommand{\XXh}{\begin{bmatrix} \mX\T \ {\mX'}\T \end{bmatrix}}
\newcommand{\XXv}{\begin{bmatrix} \mX \\ \mX' \end{bmatrix}}

We want to translate our results for linear models in Section \ref{sec:linear_th} to the kernel setting by phrasing interesting quantities in terms of inner products $\vx_i\T \vx_j$.
After finishing this procedure, we will reinterpret $\vx_i\T \vx_j \rightarrow \vphi(x_i)\T \vphi(x_j) = \K(x_i,x_j)$.

It will be useful to perform some setup before diving in.
We define $\mX = [\vx_1, ..., \vx_d]\T \in \R^{n \times m}$ and $\mX' = [\vx'_1, ..., \vx'_d]\T$ and define the full dataset kernel matrix
$\tilde{\mK} = \XXv \XXh$.
Recall that $\mGamma = \frac{1}{2n} \left( \mX\T \mX' + {\mX'}\T \mX \right)$.
In what follows, all interesting vectors will lie in either the row-space or column-space of $\XXv$ (whichever is appropriate), and so all vectors over the dataset (i.e. with $2n$ elements) will lie in the cokernel of $\tilde{\mK}$, so we are free to use the pseudoinverse of $\tilde{\mK}$ without worrying about the action of null eigendirections.
All matrix inverses henceforth will be interpreted as pseudoinverses.

\textbf{Proof of clause (a).}
An eigenpair $(\gamma, \vg)$ of $\mGamma$ satisfies
\begin{equation} \label{eqn:gamma_eigenval_def}
\mGamma \vg = \gamma \vg.
\end{equation}
Assume that $\gamma \neq 0$ and $\norm{\vg} = 1$.
Let us define a vector $\vb$ dual to $\vg$ as
\begin{equation} \label{eqn:b_def}
    \vb = \tilde{\mK}^{-1/2} \XXv \vg
    \quad
    \Leftrightarrow
    \quad
    \vg = \XXh \tilde{\mK}^{-1/2} \vb.
\end{equation}
Note that $\norm{\vg} = \norm{\vb}$.
Plugging Equation \ref{eqn:b_def} into Equation \ref{eqn:gamma_eigenval_def}, we find that
\begin{equation}
\mGamma \XXh \tilde{\mK}^{-1/2} \vb = \gamma \XXh \tilde{\mK}^{-1/2} \vb.
\end{equation}
Left-multiplying both sides by $\tilde{\mK}^{-1/2} \XXv$ yields
\begin{align}
    \tilde{\mK}^{-1/2} \XXv \mGamma \XXh \tilde{\mK}^{-1/2} \vb
    &= \gamma \tilde{\mK}^{-1/2} \XXv \XXh \tilde{\mK}^{-1/2} \vb \\
    &= \gamma \vb.
\end{align}
Defining
\begin{equation}
    \mZ \equiv
    \XXv \mGamma \XXh =
    \frac{1}{2n}
\begin{bmatrix}
    \mX {\mX'}\T \mX {\mX}\T & \mX {\mX'}\T \mX {\mX'}\T \\
    \mX' {\mX'}\T \mX {\mX}\T & \mX' {\mX'}\T \mX {\mX'}\T
\end{bmatrix}
+
\frac{1}{2n}
\begin{bmatrix}
    \mX {\mX}\T \mX' {\mX}\T & \mX {\mX}\T \mX' {\mX'}\T \\
    \mX' {\mX}\T \mX' {\mX}\T & \mX' {\mX}\T \mX' {\mX'}\T
\end{bmatrix}
\end{equation}
as in the main text and $\mK_\mGamma \equiv \tilde{\mK}^{-1/2} \mZ \tilde{\mK}^{-1/2}$, we find that
$\gamma$ is also an eigenvalue of $\mK_\mGamma$.
The same argument run in reverse (now rewriting $\vb$ in terms of $\vg$) yields that nonzero eigenvalues of $\tilde{\mK}_\mGamma$ are also eigenvalues of $\mGamma.$ \QED

\textbf{Proof of clause (b).}
The top spectral weights given by Definition \ref{def:spectramax} yield the spectral representation
\begin{equation}
    \vf(\vx) = \mU \tilde{\mS} [\vg_1 \ ... \ \vg_d]\T \vx.
\end{equation}
Plugging in Equation \ref{eqn:b_def} yields that
\begin{equation}
   \vf(\vx) = \mU \tilde{\mS} [\vb_1 \ ... \ \vb_d]\T \tilde{\mK}^{-1/2} \XXv \vx
\end{equation}
as claimed by Proposition \ref{prop:kernelized_sol}. \QED

\textbf{Proof of clause (c).}
The RKHS norm of a function with respect to a kernel $\K$ is equivalent to the minimal Frobenius norm of an explicit matrix transformation on the hidden features of the kernel, so this clause follows from Proposition \ref{prop:Fnorm_min}. \QED

\textbf{Kernelized dynamics.}

Via direct kernelization of Result \ref{res:small_init}, the kernelized representations trained from small init at an arbitrary finite time $t$ are well-approximated by
\begin{equation} \label{eqn:kernelized_traj}
    \vf(\vx, t)
    = \mU \tilde{\mS}(t) [\vg_1 \ ... \ \vg_d]\T \vx
    = \mU \tilde{\mS}(t) [\vb_1 \ ... \ \vb_d]\T \tilde{\mK}^{-1/2} [\mK_{x \X} \ \mK_{x \X'}]\T,
\end{equation}
with the singular values on the diagonal of $\tilde{\mS}(t)$ evolving according to Proposition \ref{prop:exact_dynamics}.
The initialization-dependent matrix $\mU$ and effective initial singular values are found in the explicit linear case as
\begin{equation}
    \mU \tilde{\mS}(0) = \mathcal{A}(\mW(0) {\mGamma^{(\le d)}}\T) = \mathcal{A}(\mW(0) [\vg_1, ..., \vg_d]).
\end{equation}
Using Equation \ref{eqn:b_def}, we then have that
\begin{equation} \label{eqn:kernelized_init}
    \mU \tilde{\mS}(0) = \mathcal{A}
    \left(
    [\vf(x_1), ..., \vf(x_n), \vf(x'_1), ..., \vf(x'_n)] \tilde{\mK}^{-1/2} [\vb_1, ... \vb_d]
    \right).
\end{equation}
Equations \ref{eqn:kernelized_traj} and \ref{eqn:kernelized_init} together permit one to find the trajectories from arbitrary small init in fully kernelized form.

\section{Derivation of dynamics from generic small initialization}
\label{app:derivation}

Here we give an informal derivation of Result \ref{res:small_init}, which states that the solution of Equation \ref{eqn:dW_dt_expanded} from arbitrary initialization with scale $\alpha \ll 1$ closely matches the analytical solution from a certain spectrally aligned initialization.
Recall that our ingredients are the following:
\begin{itemize}
\item $\gamma_1, ..., \gamma_d$ are unique and positive.
\item $\tilde{\mW}_0 \in \R^{d \times m}$ with $\tilde{\mW}_0 \mGamma^{(\le d)}$ full rank.
\item $\mW(t)$ is the true solution with $\mW(0) = \alpha \tilde\mW_0$.
\item $\mW^*(t)$ is the spectrally aligned solution with $\mW^*(0) = \mathcal{A}(\alpha \tilde\mW_0)$ whose dynamics are given exactly by Proposition \ref{prop:exact_dynamics}.
\end{itemize}
We will show that, for sufficiently small $\alpha$, the true and aligned solutions remain arbitrarily close.

We will find it convenient to parameterize $\mW(t)$ as
\begin{align} \label{eqn:generic_W_fac}
    \mW(t) &=
    \mU
    \begin{bmatrix} s_1(t)  & a_{1 2 }(t)  & a_{1 3 }(t)  & \cdots  & a _{1 d }(t)  & \cdots  & a _{1 m }(t)  \\ a_{21}(t)  & s_2(t)  & a _{2 3 }(t)  & \cdots  & a _{2 d }(t)  & \cdots  & a _{2 m }(t)  \\ a_{31}(t)  & a_{32}(t)  & s_3(t)  & \cdots  & a _{3 d }(t)  & \cdots  & a _{3 m }(t)  \\ \vdots  & \vdots  & \vdots  & {\ddots}  & \vdots  &   & \vdots  \\ a_{d1}(t)  & a_{d2}(t) & a_{d3}(t) & \cdots  & s_d(t)  & \cdots  & a_{dm}(t)  \\  \end{bmatrix}
    \mGamma^{(\le m)} \\
    &= \mU \mA(t) \mGamma^{(\le m)},
\end{align}
where $s_j(0) > 0$ and $a_{jk}(0) = 0$ for all $j > k$.
We use the special notation $s_j$ for the diagonal elements of $\mA$ to foreshadow that these will act as the effective singular values of the dynamics.
Note that the spectrally aligned initialization $\mathcal{A}(\tilde\mW_0)$ is precisely the $\mW(0)$ of Equation \ref{eqn:generic_W_fac} but with all off-diagonal entries of $\mA(0)$ zeroed out.
Our strategy will be to show that no $a_{jk}(t)$ ever grows sufficiently large to affect the dynamics to leading order, and thus $\mW(t)$ and $\mW^*(t)$ remain close.

We will make use of big-$\O$ notation to describe the scaling of certain values with $\alpha$.
Eigenvalues $\gamma_j$ and differences $\gamma_j - \gamma_{j+1}$ will be treated as constants.
Note that, because $\mW(0) = \alpha \tilde\mW_0$, all elements of $\mA(0)$ are $\Theta(\alpha)$ if they are not zero.

\textit{Diagonalization of dynamics.}
Define $\mLambda = \text{diag}(\gamma_1, ..., \gamma_m)$.
The dynamics of Equation \ref{eqn:dW_dt_expanded} state that $\mA(t)$ evolves as
\begin{equation} \label{eqn:dA_dt_exact}
    \frac{d\mA(t)}{dt} = \left( \mI - \mA(t) \mLambda \mA\T(t) \right) \mA(t) \mLambda,
\end{equation}
where we have reparameterized $t \rightarrow t/4$ to absorb the superfluous prefactor of $4$.

\textit{Approximate solution to dynamics.}
So long as all $a_{jk}$ remain small (i.e. $o(1)$), then these dynamics are given by
\begin{equation} \label{eqn:dA_dt_approx}
    \frac{d\mA(t)}{dt} \approx \left( \mI - \fourdiag{\gamma_1 s_1^2(t)}{\gamma_2 s_2^2(t)}{\gamma_3 s_3^2(t)}{\gamma_d s_d^2(t)} \right) \mA(t) \mLambda.
\end{equation}
We will show that all $a_{jk}$ indeed remain small under the evolution of Equation \ref{eqn:dA_dt_approx}, and so Equation \ref{eqn:dA_dt_approx} remains valid.

Solving Equation \ref{eqn:dA_dt_approx} yields
\begin{align}
    \label{eqn:s_j_ideal}
    s_j(t) &= \frac{e^{\gamma_j t}}{\sqrt{s_j^{-2}(0) + (e^{2 \gamma_j t} - 1) \gamma_j}} \\
    \label{eqn:a_jk_ideal}
    a_{jk}(t) &= 
    \left\{\begin{array}{ll}
        a_{jk}(0)
        \left( \frac{s_j(t)}{s_j(0)} \right)^{\gamma_k / \gamma_j}
        = \O\left( \alpha^{1 - \gamma_k / \gamma_j} \right)
        & \text{for } j < k, \\
        0 & \text{for } j > k.
        \end{array}\right.
\end{align}
As discussed in the main text, each $s_j(t)$ remains small up to a time $\tau_j \sim - \gamma_j^{-1} \log{\alpha}$, at which it quickly grows until $\gamma_j s_j^2(t) = 1$ and saturates.
Entries of $\mA(t)$ below the diagonal remain zero.
Entries of $\mA(t)$ above the diagonal exhibit interesting dynamics: entry $a_{jk}(t)$ with $j < k$ grows exponentially at rate $\gamma_k$, but its growth is curtailed by the saturation of $s_j(t)$ before it has time to reach order one.
This is because each $s_j(t)$ grows faster than all $a_{jk}(t)$ in its row.

All $a_{jk}(t)$ thus remain $o(1)$ and all $s_j(t)$ closely follow the ideal solution of Equation \ref{eqn:s_j(t)}, and so $\norm{\mW(t) - \mW^*(t)}_F$ remains $o(1)$.
This concludes the derivation.

\begin{figure}
  \centering
  \includegraphics[width=11cm]{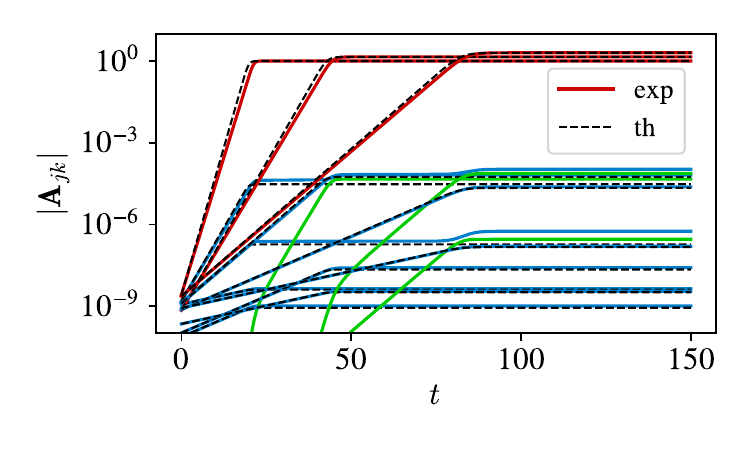}
  \vspace{-3mm}
  \caption{
  True $|s_j(t)|$ and $|a_{jk}(t)|$ compared with theoretical predictions from small init.
  Red traces show $s_j(t)$, blue traces show $a_{jk}(t)$ with $j < k$, and green traces show $a_{jk}$ with $j > k$.
  While there are no theoretical traces for $a_{jk}$ with $j > k$, these elements do remain small as predicted.
  }
  \label{fig:A_jk_traces}
\end{figure}

\textit{Numerical validation of approximation.}
While the numerical experiment presented in Figure \ref{fig:banner_figure} validates our claim regarding the trajectory of $\mW(t)$ from generic small initialization closely matching theoretical predictions from aligned initialization, here we go a step further and show agreement for individual elements of $\mA(t)$.
We analytically solve Equation \ref{eqn:dA_dt_approx} for $d = 3$ and $m = 5$ with $\gamma_j = 2^{-j}$, starting from a random upper-triangular $\mA(0)$ of scale $\alpha = 10^{-9}$.
The results, plotted in Figure \ref{fig:A_jk_traces}, closely match the unapproximated dynamics of Equations \ref{eqn:s_j_ideal} and \ref{eqn:a_jk_ideal}.

\section{Experimental details}
\label{app:exp}

Below we describe our suite of SSL experiments.
All code and experiment configs are available at \href{https://gitlab.com/generally-intelligent/ssl_dynamics}{\url{https://gitlab.com/generally-intelligent/ssl_dynamics}}.

\subsection{Single-hidden-layer MLP}
This experiment follows the linear model numerical experiment described in Section \ref{subsec:linear_model_sim}. We train a single-hidden-layer MLP for 7000 epochs over a fixed batch of 50 images from CIFAR10 using full-batch SGD. Each image is subject to a random 20x20 crop and no other augmentations. The learning rate is $\eta=0.0001$ and weights are scaled upon initialization by $\alpha=0.0001$. The hidden layer has width 2048 and the network output dimension is $d = 10$. We use Barlow Twins loss, but do not apply batch norm to the embeddings when calculating the cross-correlation matrix. $\lambda$ is set to 1.

\subsection{Full-size models}
We run two sets of experiments experiments with ResNet-50 encoders which respectively use \textit{standard init} and \textit{small init} whose results appear in figures in this paper. The standard set aims to reproduce performance reported in source publications and match original hyperparameters, whereas the small-init set includes modifications which aim to bring out the stepwise learning dynamics predicted by our theory more clearly.
Relative to the standard set, the small init set multiplies all initial weights by a small scale factor $\alpha$ and uses a reduced learning rate $\eta$.
The parameters used for each method are generally identical across the two sets except for $\alpha$ and $\eta$.

For all experiments in the standard set, we keep parameters as close as possible to those in the source publication but make minor adjustments where we find they improve evaluation metrics. Additionally, in order to avoid a steep compute budget for reproduction, we train our models on STL-10 instead of ImageNet, and reduce batch size, projector layer widths and training epochs so that the models can be trained using a single GPU in under 24 hours. \textbf{Augmentations, optimizer parameters such as momentum and weight decay, and learning rate schedules are generally left unchanged from the source publications.} Below we describe deviations from the stock hyperparameters.

\textbf{Barlow Twins.}
We set batch size to 64, and update the projector to use single hidden layer of width 2048 and embedding size $d = 512$. We use the LARS optimizer \citep{you:2017-LARS} with stock learning rates. In the small-init case, we additionally set $\lambda=1$ and $\alpha=0.093$ and remove the pre-loss batch norm. In the standard case, we keep $\lambda = 0.0051$ as suggested by \citeauthor{zbontar:2021-barlow-twins} We train the networks for 320 epochs.

\textbf{SimCLR.}
We set batch size to 256, the cross-entropy temperature $\tau=0.2$ and $\eta=0.5$. We train using SGD with weight decay $1*10^{-5}$ and no LR scheduling. In the small-init case, we use $\eta=0.01$ and additionally set $\alpha=0.087$. We train the networks for 320 epochs.%
\footnote{We found that SimCLR showed noticeably more apparent stepwise behavior in the \textit{standard} case than did Barlow Twins or VICReg (see Figure \ref{fig:dynamics_realistic}), with steps visible even in the loss curve (not shown). We are rather surprised by this difference and invite attempts to replicate it.}

\textbf{VICReg.}
We set batch size to 256, reduce the projector's hidden layer width to 512 and set $d=128$. We use $\eta=0.5$ and weight decay $1*10^{-5}$ and reduce the warmup epochs to 1. We use the LARS optimizer. In the small-init case, we use $lr=0.25$ and additionally set $\alpha=0.133$. We train the networks for 500 epochs.

\textbf{Values of $\alpha$.}
The init scales $\alpha$ quoted above were tuned so as to give clear stepwise behavior while also not slowing training too much.
We found that a useful heuristic is to choose $\alpha$ such that the top eigenvalue at init is roughly $10^{-3}$ for Barlow Twins and VICReg and $10^{-6}$ for SimCLR.

\subsection{Evaluating performance}

\begin{figure}
  \includegraphics[width=17cm]{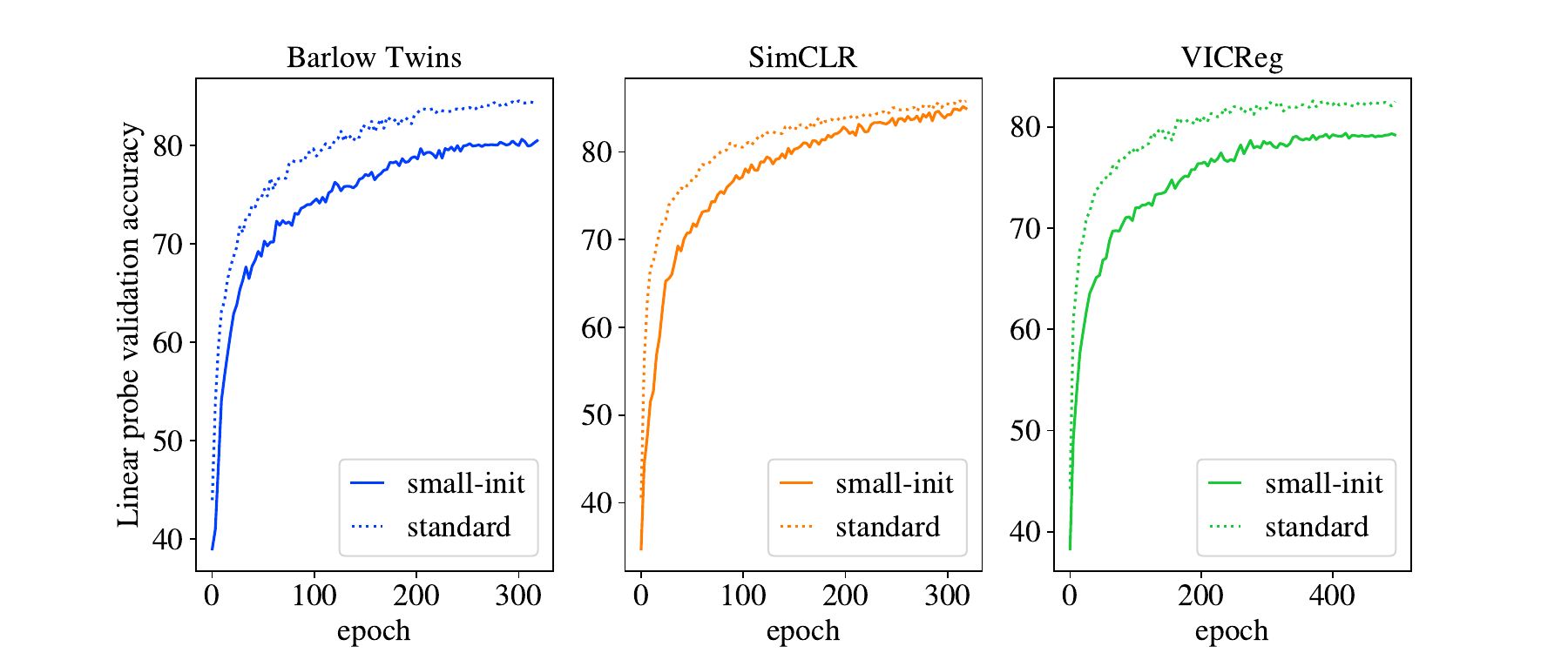}
  \vspace{-3mm}
  \caption{
    Validation accuracy of a linear probe throughout training for our standard and small-init experiments.
  }
  \label{fig:val_acc}
\end{figure}

Our small init experiments cleanly display stepwise learning.
Here we show that these hyperparameter changes do not dramatically worsen model performance.
We evaluate performance by measuring the quality of learned hidden representations using a linear probe, and perform this evaluation many times throughout training.
Validation accuracies increase throughout training (Figure \ref{fig:val_acc}) as the learned representations improve over time, and generally small-init experiments fare similarly to the standard set, with a small performance drop of a few percent, but train slower as expected.

\subsection{Measuring eigenvalues}
In the main text, all eigenvalues reported for \textit{embeddings} using the \textit{Barlow Twins loss} are the eigenvalues $\lambda_j(\mC)$, with $\mC$ the cross-correlation matrix across positive pairs as defined in Section \ref{sec:preliminaries}.
Eigenvalues reported for the embeddings of other losses are the eigenvalues $\lambda(\tilde{\mC})$ of the (centered) covariance matrix, with $\tilde{\mC} \equiv \mathbb{E}_x[(\vf(x) - \bar{\vf}) (\vf(x) - \bar{\vf})\T]$ and $\bar{\vf} \equiv \mathbb{E}_x[\vf(x)]$.
The eigenvalues reported in Figure \ref{fig:rep_vals} for hidden representations are all those of the covariance matrix (but of course computed on the hidden representations instead of the embeddings $\vf$).
We vary the matrix studied because our theory deals explicitly with $\mC$, but this cross-correlation matrix is not necessarily meaningful except in the embeddings of Barlow Twins.
In practice, however, we see similar results for either case (and in fact sometimes even see sharper learning steps in $\lambda_j(\tilde{\mC})$ than $\lambda_j(\mC)$ even for Barlow Twins).

\subsection{Measuring the empirical NTK}
In order to calculate the empirical NTK (eNTK), we employ functorch \citep{he:2021-functorch} and express the eNTK as a Jacobian contraction. We find that calculating the full kernel for realistic-sized networks is unfeasible due to memory and runtime constraints, primarily because of the large output dimension $d$ (which is 50 in the experiments of Appendix \ref{app:emb_pred}) as well as large batch sizes. To overcome this, we employ two tricks.
First, we reduce the model output to a scalar by taking an average of the outputs and calculate the eNTK for this modified function instead.\footnote{More precisely, we take the average of the $d$ outputs and multiply it by $\sqrt{d}$, a scaling which recovers the original NTK matrix in the ideal, infinite-width case.}\footnote{Note that a simplification like this is necessary to use our theory: our theory assumes the NTK on $n$ samples is captured by a single matrix, whereas in reality it is a \textit{rank-four tensor} for a model with vector output.} Second, we chunk up the computation for the full batch into minibatches depending on the size of $d$ and available GPU memory. With this approach, computing the eNTK on 1000 image pairs takes on the order of 1 minute on a single consumer GPU.
\section{Predictions of embeddings from the NTK after training}
\label{app:emb_pred}

Here we demonstrate that the true embeddings learned by SSL methods using small initialization show fair agreement with predictions computed from our theory using the empirical NTK after training.
The motivation for this experiment comes from an observation of \citet{atanasov:2021-silent-alignment} regarding ResNets with small initialization trained in a supervised setting.
Though the NTK of such a model evolves dramatically during training, the final function is nonetheless well-predicted by kernel regression using the empirical NTK after training.
We find a similar result here.

We train ResNet-50 encoders from moderately small init to close to convergence with Barlow Twins ($\alpha=0.542$), SimCLR ($\alpha=0.284$), and VICReg ($\alpha=0.604$) losses on STL-10.
These models have only $d = 50$, which is large enough to be nontrivial but small enough to give good statistical power to our subsequent analysis.
After training, we then take a batch of $n = 1000$ random augmented image pairs $\X$ and compute both their joint empirical NTK $\tilde{\mK} \in \R^{2n \times 2n}$ and their embeddings $\vf_\text{exp}(\X) \in \R^{d \times 2n}$.
We trust that $n$ is sufficiently larger than $d$ that it is reasonable to treat $\X$ as the full population distribution.
We then compute the theoretically predicted embeddings using Equation \ref{eqn:kernelized_top_spectral_embs} as
\begin{equation}
    \vf_\text{th}(\X) = \tilde{\mS} [\vb_1 \ ... \ \vb_d]\T \tilde{\mK}^{1/2}.
\end{equation}

It is principally the subspace (in function space) spanned by a vector representation which determines the performance of a linear probe on a downstream task.
As such, we compute the right singular vectors of $\vf_\text{exp}(\X)$ and $\vf_\text{th}$, which we call $\mP_\text{exp} \in \R^{d \times 2n}$ and $\mP_\text{th} \in \R^{d \times 2n}$ and which both have rank $d$.
We then compute the \textit{normalized subspace alignment} $\a(\mP_\text{exp}, \mP_\text{th})$, where $\a(\mP, \mP') \equiv \frac{1}{d} \norm{\mP (\mP')\T}_F^2$.
This alignment metric attains its maximal value of $d$ when both subspaces contain the same vectors, and has an expectation of $\frac{d^2}{2 n} \ll 1$ for random subspaces.\footnote{One may intuitively think of $\a(\mP, \mP')$ as the mean fraction of a random vector from the rowspace of $\mP$ which is captured by the rowspace of $\mP'$.}
As an additional test, we repeated this comparison replacing $\mP_\text{th}$ with $\mP_\text{NTK}$ containing the top $d$ eigenvectors of $\tilde{\mK}$ and found similar alignment.

\begin{table}
\begin{centering}
\begin{tabular}{ccccc}
  & BT & SimCLR & VICReg & (random subspaces)\\ \hline
$\a(\mP_\text{exp}, \mP_\text{th})$ & 0.615 & 0.517 & 0.592 & 0.025 \\
$\a(\mP_\text{exp}, \mP_\text{NTK})$ & 0.510 & 0.450 & 0.481 & 0.025
\end{tabular}
\caption{
\label{tab:th-exp-emb-alignment}
Alignments between true embedding subspaces and those predicted from the final NTK for different SSL methods.
}
\end{centering}
\end{table}

\begin{table}
\begin{centering}
\begin{tabular}{cccc}
$\a(\mP^\text{BT}_\text{exp}, \mP^\text{SC}_\text{exp})$ &
$\a(\mP^\text{BT}_\text{exp}, \mP^\text{VR}_\text{exp})$ &
$\a(\mP^\text{SC}_\text{exp}, \mP^\text{VR}_\text{exp})$ & (random subspaces)
\\ \hline
0.504 & 0.504 & 0.405 & 0.025\\
\end{tabular}
\caption{
\label{tab:cross-method-alignment}
Alignments between true embedding subspaces for different SSL methods.
}
\end{centering}
\end{table}

We report our observed subspace alignments in Table \ref{tab:th-exp-emb-alignment}.
For all three methods, we see an alignment between 0.5 and 0.6, which is significantly higher than expected by chance, but still misses roughly half of the span of the true embeddings\footnote{Agreement with predictions from the \textit{initial} NTK (not reported) is generally worse but still greater than chance.}.
We anticipate that much of the gap from unity is due to approximation error caused by taking a finite dataset of size $2n$.
It is perhaps surprising that our theory seems to work equally well for SimCLR and VICReg as it does for Barlow Twins.

We also report alignment scores \textit{between $\mP_\text{exp}$ from the three SSL methods} in Table \ref{tab:cross-method-alignment}, again finding alignment roughly between 0.4 and 0.5.
This finding, independent of any theoretical result, is evidence that these three methods are in some respects learning very similar things.

\section{Connection to spontaneous symmetry breaking in physical systems}
\label{app:symm}

In the main text, we noted that \citet{landau:1944-turbulence} encountered essentially the same dynamics in the study of turbulent fluid flow as we found for SSL.
This is due to the fact that both are simple processes of \textit{spontaneous symmetry breaking} (SSB), a phenomenon in which a system whose dynamics obey a symmetry spontaneously settles into one of several asymmetric rest states chosen as a result of the system's initial conditions.
In this appendix, we will explain how the dynamics of our model of SSL can be understood as a process of SSB.

Recall from Equation \ref{eqn:L_as_eigensum} that the loss of our toy model upon aligned initialization is
\begin{equation}
    \L = \sum_j \L_j = \sum_j (1 - \gamma_j s_j^2)^2 = \sum_j (1 - \bar{s}_j^2)^2,
\end{equation}
where we define $\L_j = (1 - \gamma_j s_j^2)^2$ and $\bar{s}_j = \gamma_j^{1/2} s_j$.
The quantity $\bar{s}_j$ is a (rescaled) singular value of $\mW$.
Singular values are canonically taken to be nonnegative (because one can always enforce this condition by appropriate choice of the singular vectors), but let us pretend for the present discussion that singular values may take any value along the real line.
Note that each singular value evolves via gradient descent according to its respective loss as
\begin{equation}
    \bar{s}_j'(t) = \frac{d \L_j}{d \bar{s}_j(t)}.
\end{equation}

\begin{figure}
  \centering
  \includegraphics[width=12cm]{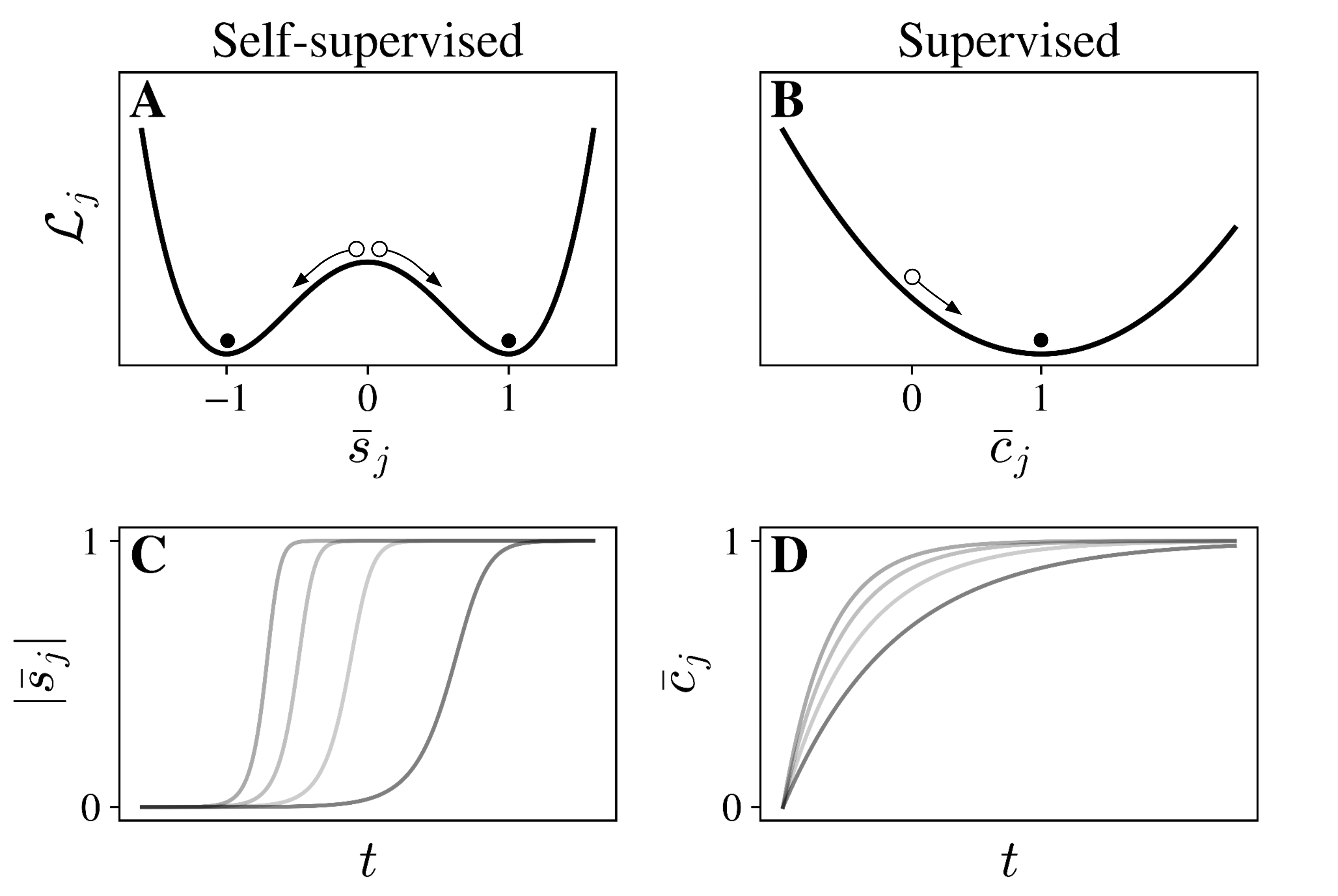}
  \vspace{-3mm}
  \caption{
    \textbf{SSL is a process of symmetry breaking, whereas standard supervised learning is not.}
    \textbf{(A)}: Nondimensionalized loss landscape for singular value $\bar{s}_j$ in our linear model of Barlow Twins.
    The system can reach one of two minima depending on the sign of the initialization.
    \textbf{(B)}: Nondimensionalized loss landscape for an eigencoefficient in linearized \textit{supervised} learning with $\bar{c_j}^* = 1$.
    There is only one local minimum.
    \textbf{(C)}: Example trajectories for the loss landscape in (A) from small init with different eigenvalues $\gamma_j$.
    \textbf{(D)}: Example trajectories for the loss landscape in (B) from small init with different eigenvalues $\gamma_j$.
  }
  \label{fig:ssb_cartoon}
\end{figure}

Figure \ref{fig:ssb_cartoon}A depicts $\L_j$, with trajectories for various $\gamma_j$ shown in Figure \ref{fig:ssb_cartoon}C.
Note that this 1D landscape is symmetric and has two global minima at $\bar{s}_j = \pm 1$ and one local maximum at $\bar{s}_j = 0$.
When the system is initialized near zero, an unstable equilibrium of the dynamics, it will flow towards one basin or the other depending on its initial sign.
However, for small $\bar{s}_j(0)$, it takes a time $\tau_j = -\log(|\bar{s}_j(0)|) / 4 \gamma_j$ for $\bar{s}_j$ to escape the origin,\footnote{To be more precise, it takes a time $\tau_j^{(\epsilon)} \approx -\log(\epsilon / |\bar{s}_j(0)|) / 4 \gamma_j$ to escape a ball of small constant radius $\epsilon > 0$ around the origin, and we drop the sub-leading-order contribution of $\epsilon$.} whereas the system thereafter approaches the nearest minimum with the faster timescale $\tau'_j = 1/4\gamma_j$.
This slow escape from the origin is what leads to the sharp stepwise behavior we find.

It should be noted that the quartic form of $\L_j$ appears in toy models of SSB across physics (see e.g. Landau-Ginzburg theory \citep{kardar:2007-stat-phys-of-fields}) and is in fact the aforementioned model of \citet{landau:1944-turbulence}.
In these and other typical cases, SSB entails the breaking of a symmetry apparent at the outset of the problem such as invariance to translation, rotation, or inversion symmetry.
In the case of SSL, the symmetry is the transformation $\vf \rightarrow - \vf$, which does not change the value of the global loss.\footnote{Our model in fact obeys the more general symmetry $\vf \rightarrow \mU \vf$ for any orthonormal matrix $\mU$, which is shared by SimCLR but not by VICReg or Barlow Twins with $\lambda \neq 1$.}

Why does standard supervised learning in the NTK limit not exhibit stepwise behavior upon small initialization?
Following the analysis of \citet{jacot:2018}, the analogous modewise loss for a supervised setup takes the form $\L_j \propto (\bar{c}^*_j - \gamma_j c_j)^2$, with $c_j$ a learnable coefficient of Gram matrix eigenvector $j$ in the representer theorem coefficients of the learned function and $\bar{c}^*_j$ a constant.
We nondimensionalize in this case as $\bar{c}_j = \gamma_j c_j$.
As shown in Figure \ref{fig:ssb_cartoon}B, the landscape in the supervised case is merely quadratic and has no unstable equilibria, reflecting the lack of inversion symmetry in the loss.
Therefore $\bar{c}_j'(0)$ is not small, and the respective coefficients grow immediately rather than first undergoing a period of slow growth, as shown in Figure \ref{fig:ssb_cartoon}D.

The main surprising finding of our experiments is that SSL experiments with ResNets exhibit stepwise learning (especially upon small init) even far from the linear (i.e. lazy) regime.
It thus seems likely that one could derive the stepwise phenomenon from a much looser set of assumptions on the model class.
The view of stepwise learning as a simple consequence of SSB may be of use in the development of a more general theory in this sense.
This SSB view suggests that we may understand the growth of each new embedding direction from zero as the escape from a saddle point and use the fact that the local loss landscape around any saddle point of a given index (and with nondegenerate Hessian) is the same as any other up to rescaling of the domain.
We conjecture that stepwise learning will occur generically modulo edge cases (such as simultaneous growth of multiple directions) given an appropriate choice of loss function under minimal and realistic conditions on the empirical (i.e. time-varying) NTK of the model and assuming the final embeddings are the output of a linear readout layer.

\section{Potential modifications for speeding up SSL}
\label{app:speedup}

Compared to standard supervised learning, SSL is known in folklore to be much slower to train.
Our work presents a theory of the training dynamics of SSL, and thus it is natural to ask whether it sheds light on this point of difficulty or suggests means by which it might be addressed.
Here we suggest an explanation for the slowness of SSL training and propose various fixes which are ripe for future study.

In our picture of the training dynamics of SSL, embedding eigenvalues start small and grow sequentially, converging when $d$ eigenvalues have sufficiently grown.
Smaller eigendirections require more time to grow.
\textbf{We suggest that SSL is slow to converge because one must wait for the small eigendirections to grow.}
This hypothesis is supported by Figure \ref{fig:dynamics_realistic}, which shows that, in a realistic configuration, a significant fraction of the eigenvalues remain small even late in training.\footnote{This observation is particularly compelling in light of the finding of \citet{garrido:2022-rankme} that generalization is better when embeddings have higher rank. Their work suggests that hyperparameters should be chosen to maximize the embedding rank at the end of training.}

This suggests that SSL can be sped up by modifying training to target small eigenvalues.
Here we suggest one way this might be achieved via preconditioning of gradients and two additional ways this might be achieved by modifying the loss function itself.
We leave our results at the level of theoretical speculation.
We encourage interested SSL practitioners to try implementing the methods described and reach out with questions.

\subsection{Targeting gradients towards small PCA directions.}
One potentially promising idea is to simply compute the PCA matrix of the embeddings and ``manually" increase the gradient pull along directions which are very small, thereby encouraging the distribution to expand in directions in which it is currently flat.
Let us denote by $\mF \in \R^{2n \times d}$ the embeddings for a given data batch.
Backpropagation will compute $\nabla_\theta \L = \text{Tr}[ (\nabla_\theta \mF)\T \ \nabla_\mF \L ]$.
We may simply enact the substitution
\begin{equation}
    \nabla_\mF \L
    \longrightarrow
    (\nabla_\mF \L) (\mF\T \mF + \alpha \mI_d)^{-1}
\end{equation}
with $\alpha > 0$ to suppress $\nabla_\mF \L$ along directions of large variance and increase it along directions of small variance.
We anticipate this will encourage faster growth of small eigenmodes.
This trick could apply to any joint embedding method.

\begin{figure}
  \centering
  \includegraphics[width=12cm]{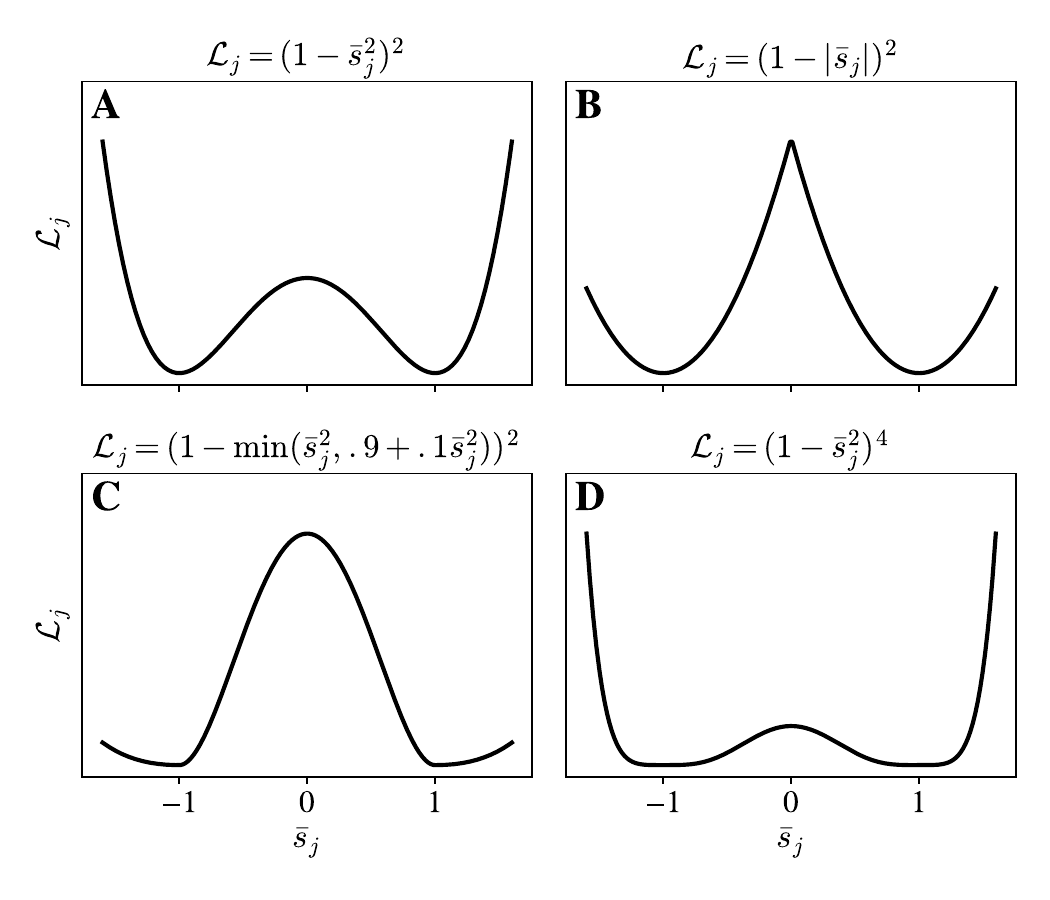}
  \vspace{-3mm}
  \caption{
    \textbf{Proposed modifications for the modewise SSL loss to encourage or enable the faster growth of slow eigendirections.}
    Losses are expressed as a function of $\bar{s}_j = \gamma_j^{1/2} s_j$ and assume $\gamma_j > 0$.
    \textbf{(A)}: Original Barlow Twins modewise loss.
    \textbf{(B)}: A modified loss with a kink at zero. The origin is no longer an unstable equilibrium and nearby points will quickly escape.
    \textbf{(C, D)}: Two losses modified so as to have smaller curvature at their minima and thereby permit larger learning rates without instability.
  }
  \label{fig:speedup_cartoon}
\end{figure}

\subsection{Sharpening at zero}
This and the following method are specialized to the Barlow Twins loss.
They will make use of visual depictions of the modewise loss landscape as described in Appendix \ref{app:symm}, and so we recommend reading that appendix first.
They will both consider transformations $g(\mC)$ of the correlation matrix $\mC$, where $g$ acts pointwise on the eigenvalues of $\mC$.

Recall that we found that the singular values of the system under aligned conditions evolve under a loss of the form $\L_j = (1 - \lambda_j)^2 = (1 - \gamma_j s_j^2)^2$.
As shown in Figure \ref{fig:speedup_cartoon}A and discussed in Appendix \ref{app:symm}, this means that, when initialized near zero, they initially feel only a small gradient and must spend a long time escaping the origin, with that duration determined by just how small they are initially.
However, this is no longer the case if we change the loss so it has a kink at zero.
For example, if we change the loss to $\L_{\text{sharp}}(\mC) = \norm{g_\text{sqrt}(\mC) - \mI_d}_F$ with $g_\text{sqrt}(\lambda) = \text{sign}(\lambda) |\lambda|^{1/2}$, then all singular values with $\gamma_j > 0$ feel a $\Theta(1)$ repulsive force from the origin regardless of how close they are.
This loss is plotted in Figure \ref{fig:speedup_cartoon}B.
Interventions of this type lead to eigenvalue growth curves of the type shown in Figure \ref{fig:ssb_cartoon}D.\footnote{In fact, the particular choice $g_\text{sqrt}(\lambda) = \text{sign}(\lambda) |\lambda|^{1/2}$ gives dynamics which can be solved exactly for aligned init just like the original Barlow Twins loss studied in the main text.}

\subsection{Smoothing around minima}

The naive way to speed up the growth of slow-growing eigenmodes is simply to increase the learning rate.
This fails because the minima of the fast-growing eigenmodes will eventually become unstable.
For example, if $d=2$ and the total loss is $\L = (1 - s_1^2)^2 + (1 - 10^{-6} s_2^2)^2$, any learning rate above $\eta = 1/4$ will cause instability in $s_1$, but $s_2$ requires a larger learning rate to grow in reasonable time.
One solution here is to modify the loss landscape so that the higher eigendirections see a wider basin around their minima and can thus tolerate a larger learning rate.
One implementation might be $\L_{\text{smooth}}(\mC) = \norm{g_\text{smooth}(\mC) - \mI_d}_F$ with $g_\text{smooth}(\lambda) = \min(\lambda, 1)$.
One might replace this with e.g. $g_\text{smooth} = \min(\lambda, 1 + \epsilon (1 - \lambda))$ for some small $\epsilon > 0$.
Alternatively, one might modify the structure of the loss function itself to be e.g. $\L_j = (1 - \lambda_j)^4$, which has vanishing curvature at its minimum, or implement a similar idea with a hinge function to create a perfectly flat basin of finite width.
Two possibilities for losses modified to have wider minima are shown in Figures \ref{fig:speedup_cartoon}C,D.

Both these last ideas are specialized to Barlow Twins in the case of $\lambda = 1$, but we anticipate they could be easily generalized.
In fact, VICReg already includes a square root in the variance term which we conjecture implicitly does something similar to our proposed sharpening modification.

\end{document}